\documentclass{article}

\usepackage[square,numbers,compress]{natbib}

\usepackage[preprint]{neurips_2025}

\usepackage[utf8]{inputenc} 
\usepackage[T1]{fontenc}

\usepackage{amsmath,amssymb,amsbsy,amsfonts,amscd,bm,url,color,latexsym}
\usepackage{amsthm}
\usepackage[colorlinks=true,
            citecolor=blue,    
            linkcolor=red,    
            urlcolor=blue 
           ]{hyperref}

\usepackage{cleveref}  
\usepackage{paralist}
\usepackage[dvipsnames]{xcolor}
\usepackage{color}
\usepackage{graphicx}
\graphicspath{{./figs/}}
\usepackage{algorithm}
\usepackage{algorithmic}
\usepackage{comment}
\usepackage{multirow}
\usepackage{enumitem}
\usepackage{fancyhdr}
\usepackage{cite}
\usepackage{subcaption}
\usepackage{authblk}
\usepackage{booktabs} 
\usepackage{wrapfig}
\usepackage{adjustbox}

\newtheorem{theorem}{Theorem}[section]

\usepackage{enumitem}
\usepackage{tabularx}

\newcommand{\R}{\mathbb{R}}

\newcommand{\e}{\begin{equation}}
\newcommand{\ee}{\end{equation}}
\newcommand{\en}{\begin{equation*}}
\newcommand{\een}{\end{equation*}}
\newcommand{\eqn}{\begin{eqnarray}}
\newcommand{\eeqn}{\end{eqnarray}}
\newcommand{\bmat}{\begin{bmatrix}}
\newcommand{\emat}{\end{bmatrix}}

\DeclareMathAlphabet\mathbfcal{OMS}{cmsy}{b}{n}

\newcommand{\E}{\operatorname{\mathbb{E}}}

\newcommand{\mc}{\mathcal}

\newcommand{\vct}[1]{\boldsymbol{#1}}
\newcommand{\mtx}[1]{\boldsymbol{#1}}

\DeclareMathOperator*{\argmin}{\text{arg~min}}

\newcommand{\NC}{$\mc {NC}$}

\newcommand{\calH}{\mathcal{H}}

\newcommand{\calS}{\mathcal{S}}

\newcommand{\calX}{\mathcal{X}}

\newcommand{\va}{\vct{a}}

\newcommand{\vh}{\vct{h}}

\newcommand{\vk}{\vct{k}}

\newcommand{\vq}{\vct{q}}

\newcommand{\vv}{\vct{v}}

\newcommand{\vmu}{\vct{\mu}}

\newcommand{\mW}{\mtx{W}}

\newcommand{\mZ}{\mtx{Z}}

\graphicspath{{./figs/}}

\newlength{\imgwidth}
\newboolean{twoColVersion}
\setboolean{twoColVersion}{false}
\newcommand{\twoCol}[2]{\ifthenelse{\boolean{twoColVersion}} {#1} {#2} }

\usepackage{tikz}
\usepackage{tikz-3dplot}
\usepackage{pgfplots}
\usepgfplotslibrary{patchplots}
\usetikzlibrary{patterns, positioning, arrows}
\pgfplotsset{compat=1.15}

\usepackage[most]{tcolorbox}
\newtcolorbox{titlebox}[1]{%
    tikznode boxed title,
    enhanced,
    arc=0mm,
    boxrule=1pt,
    halign=left,
    bottom=1ex,
    interior style={white},
    attach boxed title to top left={xshift=3ex,yshift=-\tcboxedtitleheight/2},
    fonttitle=\bfseries,
    colbacktitle=white,coltitle=black,
    boxed title style={size=small,colframe=white,boxrule=0pt},
    title={#1}
}

\usepackage{xspace}

\newcommand{\phenomenon}{\mbox{Layerwise Compression-Expression}\xspace}

\title{From Compression to Expression: \\A Layerwise Analysis of In-Context Learning}

\begin{document}



\author{\textbf{Jiachen Jiang}, \textbf{Yuxin Dong}, \textbf{Jinxin Zhou \&} \textbf{Zhihui Zhu}\thanks{The corresponding author.}}

\affil{Department of Computer Science and Engineering, \\ The Ohio State University,\\
\texttt{\{jiang.2880, dong.1357, zhou.3820, zhu.3440\}@osu.edu}}

\maketitle

\begin{abstract}
In-context learning (ICL) enables large language models (LLMs) to adapt to new tasks without weight updates by learning from demonstration sequences. While ICL shows strong empirical performance,  its internal representational mechanisms are not yet well understood. In this work, we conduct a statistical geometric analysis of ICL representations to investigate how task-specific information is captured across layers. Our analysis reveals an intriguing phenomenon, which we term {\it Layerwise Compression-Expression}: early layers progressively produce compact and discriminative representations that encode task information from the input demonstrations, while later layers express these representations to incorporate the query and generate the prediction. This phenomenon is observed consistently across diverse tasks and a range of contemporary LLM architectures. We demonstrate that it has important implications for ICL performance---improving with model size and the number of demonstrations---and for robustness in the presence of noisy examples. To further understand the effect of the compact task representation, we propose a bias-variance decomposition and provide a theoretical analysis showing how attention mechanisms contribute to reducing both variance and bias, thereby enhancing performance as the number of demonstrations increases. Our findings reveal an intriguing layerwise dynamic in ICL, highlight how structured representations emerge within LLMs, and showcase that analyzing internal representations can facilitate a deeper understanding of model behavior.
\end{abstract}

\section{Introduction}
\label{sec:intro}

 In-context learning (ICL) \citep{brown2020language, dong2022survey}  has emerged as a powerful capability of large language models(LLMs), allowing them to perform new tasks by conditioning on a few input-output examples without weight updates. Despite being trained solely for next-token prediction, LLMs exhibit strong empirical performance across a wide range of NLP tasks through this mechanism. For example, pretrained LLMs can make correct predictions based on a sequence of input-separation-output pairs that encode semantic mappings. Given the same query token, the model can make different predictions based on the task defined by the demonstrations, such as $d$ and $C$ for the following two tasks,
\begin{align}
(\underbrace{a \to b, \  b\to c,}_{\text{Next Letter}} \ c \to ? ), \quad 
( \underbrace{a \to A, \  b\to B,}_{\text{To Uppercase}} \ c \to ? )
\label{eq:example-tasks}\end{align}

 Recent research has advanced several theoretical perspectives on explaining why ICL works—viewing ICL as posterior inference \citep{xie2021explanation}, implicit meta-learning \citep{chen2021meta}, internal optimization \citep{von2023transformers} or mechanistic interpretations\citep{olsson2022context}. However, the underlying mechanism of how LLMs distinguish different tasks and use this information to guide their output remains unclear for ICL. To address this gap, we focus on the hierarchical feature learning across layers and formulate the following research question:

 \begin{center}
 \textit{How do LLMs extract and differentiate task information from shallow to deep layers \\during in-context learning?}
 \end{center}

To investigate how task-specific information evolves across layers, we conduct a statistical geometric analysis of ICL representations across multiple tasks. Specifically, we consider a set of $T$ ICL tasks, each associated with a distribution $\mc D_t$ for input-output pairs. Then for each task $t$, we randomly sample $K$ input-output pairs (also referred to as \textit{demonstrations}) from $\mc D_t$, which are combined with a query to form an ICL {\it instance}. We construct multiple such instances per task following this procedure. 
To quantify how the model compresses task information in its internal representations, we examine two key aspects: (1) how instances from the same task are clustered together, and (2) how instances from different tasks are distinguished from each other. Our analysis reveals an intriguing phenomenon, which we term the \textbf{\phenomenon}, summarized as:
\begin{titlebox}{\phenomenon Phenomenon}
\textit{LLMs exhibiting ICL capabilities organize their layers into two parts with distinct behaviors: a compression part and an expression part. The early layers, comprising the compression part, progressively produce compact and discriminative representations that capture task information from the input demonstrations. The later layers, forming the expression part, apply these compact representations to the query to generate the output.}
\end{titlebox}

Specifically, we introduce a metric called Task-Distance Normalized Variance (TDNV)\footnote{Following a similar conceptual framework to Class-Distance Normalized Variance (CDNV) \citep{galanti2021role} by viewing each task as a class.} that measures the ratio of within-task variance to between-task distance: within-task variance indicates how well the representation from the same task are compressed, while between-task distance reflects the separation from different tasks. A lower TDNV indicates that representations of the same task samples are similar and representations of different task samples are distinguishable. Thus, TDNV serves as an effective method of how well the task information is compressed. By measuring TDNV across transformer layers, we can track how the model progressively encodes and transforms task information throughout its architecture.

 \begin{figure}[t]
    \centering
    \includegraphics[width=0.75\linewidth]{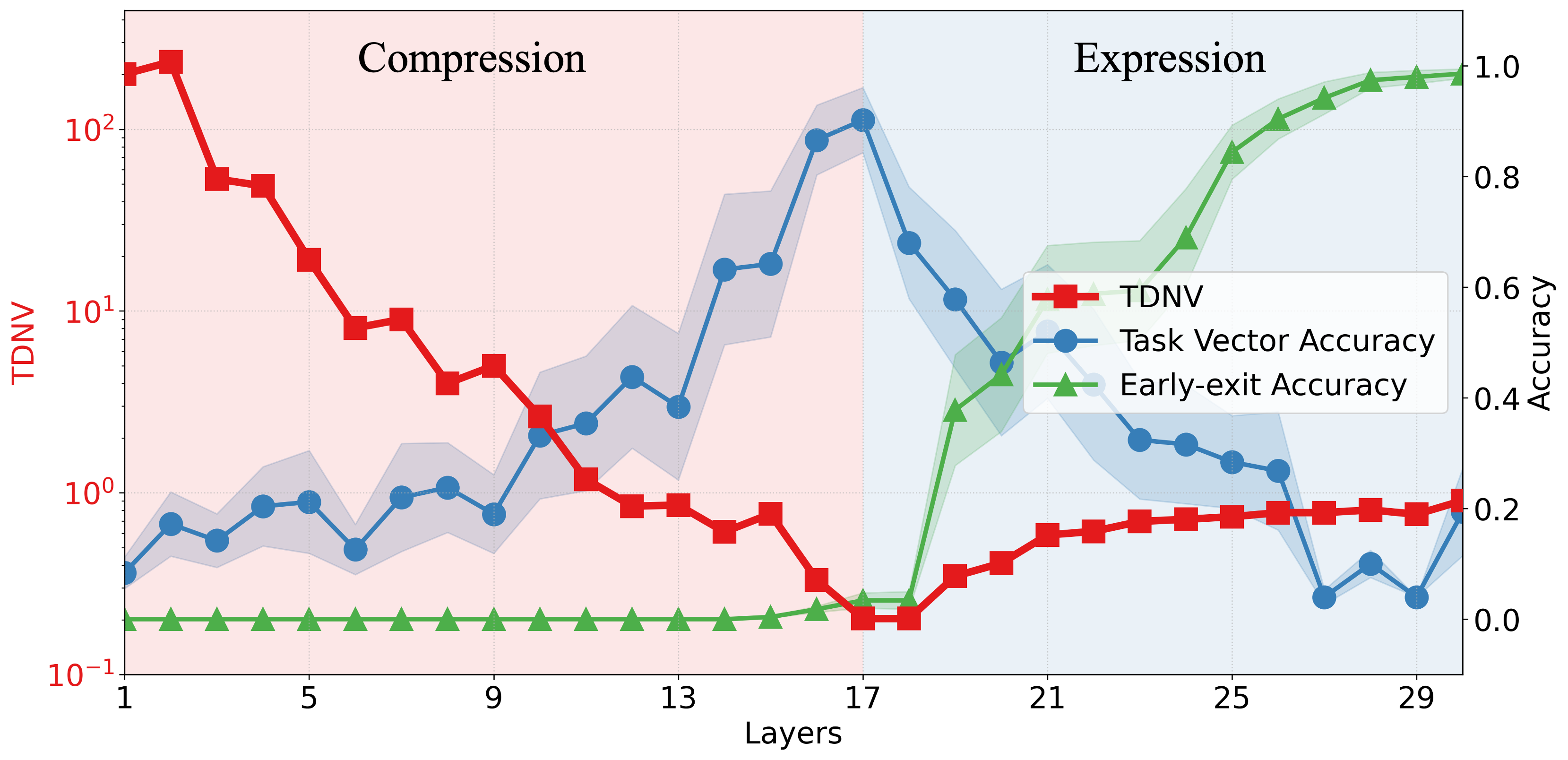}
    \caption{\textbf{Layer-wise compression to expression} in ICL representations. TDNV first decreases then increases from shallow to deep layers, splitting the model into compression and expression stages. During the compression stage, task vector accuracy increases as task information is progressively extracted from demonstration pairs. During the expression stage, early-exit accuracy increases as output information is progressively decoded based on the input query. Refer to \Cref{app:tv-early-acc} for detailed explanation of task vector and early-exit accuracy.}
    \label{fig:main}
\end{figure}

As shown in \Cref{fig:main}, TDNV first decreases and then increases from shallow to deep layers, splitting the model into compression and expression stages. To further support our hypothesis, we measure task vector accuracy \citep{hendel2023context} and early-exit accuracy \citep{xin2020deebert,jiang2024tracing} across layers to track task-specific and output-specific information. Task vector accuracy measures zero-shot ICL performance when injecting intermediate layer hidden states extracted under ICL settings. Early-exit accuracy measures performance by directly applying the final-layer classifier to intermediate hidden states. During compression, task vector accuracy increases while early-exit accuracy remains low, indicating that the model is compressing representations to encode task information. During expression, task vector accuracy decreases while early-exit accuracy rapidly increases, indicating that  the model begins to incorporate query information and decode output-specific representations. As we show, this behavior has important implications for ICL performance and robustness.

\textbf{Contributions.} Our main contributions can be summarized as follows: 
\begin{itemize}[leftmargin=12pt,itemsep=2pt,topsep=0pt,parsep=0pt]
    \item By analyzing the hidden representations of ICL, we conceptualize and extensively examine the \phenomenon phenomenon. Our results show that it is prevalent across model architectures (transformer and state-space models) and task domains (symbolic, language understanding and multimodality), and emerges during the training process. 

    \item We show that larger models and more demonstrations lead to more compressed task representations, explaining why larger models and longer contexts yield better performance. To further understand the compressed representation, we propose a bias-variance decomposition and provide a theoretical analysis showing how attention mechanisms contribute to reducing both variance and bias, thereby enhancing performance as the number of demonstrations increases. 
    \item We show that noisy demonstrations result in less compressed representations and a corresponding drop in performance. However, the representations remain distinguishable with a certain amount of noise, which helps explain the robustness of ICL. Moreover, we find that errors in early demonstrations can be suppressed by later examples, and that errors in later demonstrations lead to less compressed representations than those in early ones. This highlights the recency effect \citep{kossen2024context,yu2024large} and the key role of later demonstrations.

    \item Motivated by our analysis, we propose task-vector contrastive fine-tuning method to further compress task vectors and reduce TDNV. Fine-tuning GPT-2 models on symbolic ICL tasks with this approach yields 20\% improvement on average in task-vector accuracy over standard finetuning.
    
\end{itemize}

\textbf{Significance of the Finding.} Our analysis provides a new perspective on why {\it decoder-only} LLMs trained for next-token prediction can serve as flexible architectures for a wide range of tasks. Despite lacking an explicit bottleneck layer, these models exhibit behavior reminiscent of {\it encoder-decoder} architectures: early layers distill task information from demonstrations into compact representations, while later layers decode these representations into query-specific outputs. The compression stage aligns with the Information Bottleneck (IB) principle \citep{saxe2019information, kawaguchi2023does}, which posits that efficient neural representations are achieved by compressing inputs to retain only task-relevant information while discarding redundant or irrelevant details. However, standard IB theory focuses exclusively on the compression phase and is primarily developed in the context of classification problems. 
Our work also provides justification for previous pruning studies \citep{men2024shortgpt,luo2025adaptive}, which show that deeper layers tend to be more redundant and can be safely skipped, whereas skipping earlier layers often results in significant performance degradation.
\section{Related Works}
\label{sec:related}

\textbf{Layerwise Representations.} Prior works \citep{ben2022nearest, fang2021exploring, wang2023understanding, rangamani2023feature, he2024law, zhou2025all} investigated the role of different layers in feature learning. They revealed that in classification task, intermediate layer features become increasingly linearly separable and exhibit Neural Collapse (\NC), indicating \textbf{monotonic} feature compression with depth. In contrast, we hypothesize that decoder-only ICL models follow a \textbf{dual} process where shallow layers compress information and deeper layers re-express it, and intermediate layers achieve maximal compression.

\textbf{In-Context Learning Interpretability.} Numerous studies have investigated the mechanisms underlying ICL \citep{xie2021explanation, chen2021meta, von2023transformers, dai2022can, ahn2023transformers, olsson2022context, dong2025understanding}, spanning perspectives of Bayesian inference, meta-learning, and optimization. Our work instead analyzes through layer-wise representations. In addition, \citet{doimo2024representation} examine the geometry of ICL representations by clustering intermediate features according to semantic subjects of input, whereas our findings differ by focusing on the underlying tasks induced by input–output pairs.

\textbf{Task Representations.} Various compact representations capture ICL tasks, including task vector~\citep{hendel2023context}, function vector~\citep{todd2023function}, and state vector~\citep{li2024context}, which guide model behavior by injecting hidden states. Other works explore compositional and latent-space manipulation \citep{shao2023compositional, liu2023context}. Prior studies focus on single-task representations, whereas we provide a layer-wise geometric analysis of representations.

A more comprehensive discussion of the related works can be found in \Cref{app:related_works}.

\section{Preliminaries}
In this section, we first formally set up the layer-wise representations of in context learning in \Cref{subsec:setup}, followed by introducing the metrics for measuring within-task  compression of features at each layer in \Cref{subsec:metric}. 

\subsection{Problem Setup}
\label{subsec:setup}
\textbf{Layerwise ICL Representations.} For ICL task, we are given $(i)$ $K$ demonstrations, denoted as $\mathcal{S}_K = \{ s_1, s_2, \dots, s_K \}$, where each demonstration $s_k = (x_k \rightarrow y_k)$ consists of an input token $x_k$, a separator token "$\rightarrow$", and a label token $y_k$; and $(ii)$ a query input $\mathcal{X} = (x_q \rightarrow)$. We refer to the demonstration-query pair $(\calS_K,\calX)$ as an ICL {\it instance}. An LLM $f$ performs ICL by processing the instance $(\calS_K,\calX)$ as a next-token prediction task. Let $\mZ^{(\ell)}\in \R^{d\times p}$ denote the hidden representations at layer $\ell$ for the  instance, where $p$ denotes the sequence length and $d$ represents the dimension of the hidden representation.
The layerwise update with $f$ is performed as
\begin{equation}
\begin{aligned}
    \mZ^{(\ell + 1)} &= f_{\theta^{(\ell)}}(\mZ^{(\ell)}), \quad \text{for } \ell = 0, 1, \dots, L-1,
\end{aligned}
\end{equation}
where $f_{\theta^{(\ell)}}: \mathbb{R}^{d \times p} \rightarrow \mathbb{R}^{d \times p}$ denotes the transformation---such as self-attention and a multi-layer perceptron in a Transformer---within the $\ell$-th layer, parameterized by $\theta^{(\ell)}$.

For autoregressive models, the final prediction is produced by applying a classifier on the representation of the last separation token in the final layer $\mZ^{(L)}$, which predicts the label $y_q$ of the query input. Since this token summarizes the entire context, we use its hidden representation as the ICL representation at layer $\ell$, denoted by $\vh^{(\ell)}$ to simplify the notation. This vector is also referred to as the {\it task vector} in \citep{hendel2023context}. Throughout the remainder of this paper, we use $\vh^{(\ell)}$ to analyze layer-wise behavior and information encoding in the ICL process.

\subsection{Metric for Representation Compression}
\label{subsec:metric}
To analyze how models distinguish between different tasks, we consider $T$ ICL tasks, each with a task-specific data distribution $\{\mathcal{D}_t\}_{t = 1}^{T}$. For each task $t$, we sample $N$ ICL instances of form $(\mathcal{S}_K, \mathcal{X})$ from $\mathcal{D}_t$. For each instance, we use the hidden representation of the last token at layer $\ell$ as the representation of the inputs, denoted as $\vh_{i,t}^{(\ell)} \in \mathbb{R}^d$ for the $i$-th instance from task $t$.

The study of feature compression and discrimination \citep{yu2020learning,papyan2020prevalence, zhu2021geometric,zhai2020complete,galanti2021role,jiang2023generalized} has recently gained significant attention in representation learning. Inspired by this line of work, we analyze how models compress task information in their internal representations by examining two key aspects.
\begin{itemize}
[leftmargin=12pt,itemsep=2pt,topsep=0pt,parsep=0pt]
    \item We quantify how samples from the same task cluster together by using the \textbf{within-task variance} 
    \begin{align}
    \text{var}_t^{(\ell)} = \frac{1}{N} \sum_{i=1}^{N} \| \vh_{i,t}^{(\ell)} - \overline{\vh}_t^{(\ell)} \|_2^2, \quad \text{where}  \quad \overline{\vh}_t^{(\ell)} = \frac{1}{N} \sum_{i=1}^{N} \vh_{i,t}^{(\ell)}.
    \end{align}
    It measures how well representations from the same task are compressed toward their task mean. Specifically, when this value decreases, it indicates that features within each task are more tightly compressed around their respective means. 
\item To quantify how effectively samples from different tasks are distinguished from each other, we use the \textbf{between-task distance} of two tasks $t$ and $t'$ as $
    \| \overline{\vh}_t^{(\ell)} - \overline{\vh}_{t'}^{(\ell)} \|_2^2$.
It measures the distance between the centers of different tasks and the features of each task become more separable as this distance increases. 
\end{itemize}
We then use a metric inspired by the class-distance normalized variance used in classification tasks \citet{galanti2021role}, referred to as Task-Distance Normalized Variance (TDNV) here to measure the ratio of within-task variance to between-task distance:
\begin{equation}
    \text{TDNV}^{(\ell)}:= \sum_{t=1}^{T} \sum_{\substack{t' = 1 \\ t' \neq t}}^{T} \frac{\text{var}_t^{(\ell)} + \text{var}_{t'}^{(\ell)}}{2 \| \overline{\vh}_t^{(\ell)} - \overline{\vh}_{t'}^{(\ell)} \|_2^2}, \quad \forall \, \ell \in [L].
    \label{eq:cdnv}
\end{equation}
The decrease of TDNV indicates more compressed and discriminated feature for the ICL task.
\section{\phenomenon\ Dynamic}
\label{sec:empirical}
In this section, we examine the dynamics of layer-wise representation under the ICL setting, a phenomenon we termed \textit{\phenomenon}. The subsequent sections validate and explore this phenomenon in detail across various conditions. Specifically, \Cref{subsec:main} demonstrates that it occurs universally across different architectures and tasks. Next, we analyze key factors influencing this phenomenon, including model size (\Cref{subsec:model_size}) and demonstration noise (\Cref{sec:robustness}).

\subsection{Prevalence of \phenomenon}
\label{subsec:main}

To validate whether \textit{\phenomenon} is a general mechanism of ICL, we evaluate it across different LLM model architectures and tasks. Unless otherwise specified, we use Deepseek-coder-7B \citep{guo2024deepseek} as our default model. For each task, we sample $N = 100$ instances, setting the default number of demonstrations to $K = 15$. 

\begin{figure}[t]
    \centering
    \begin{minipage}[t]{0.48\linewidth}
        \centering
        \includegraphics[height=4cm]{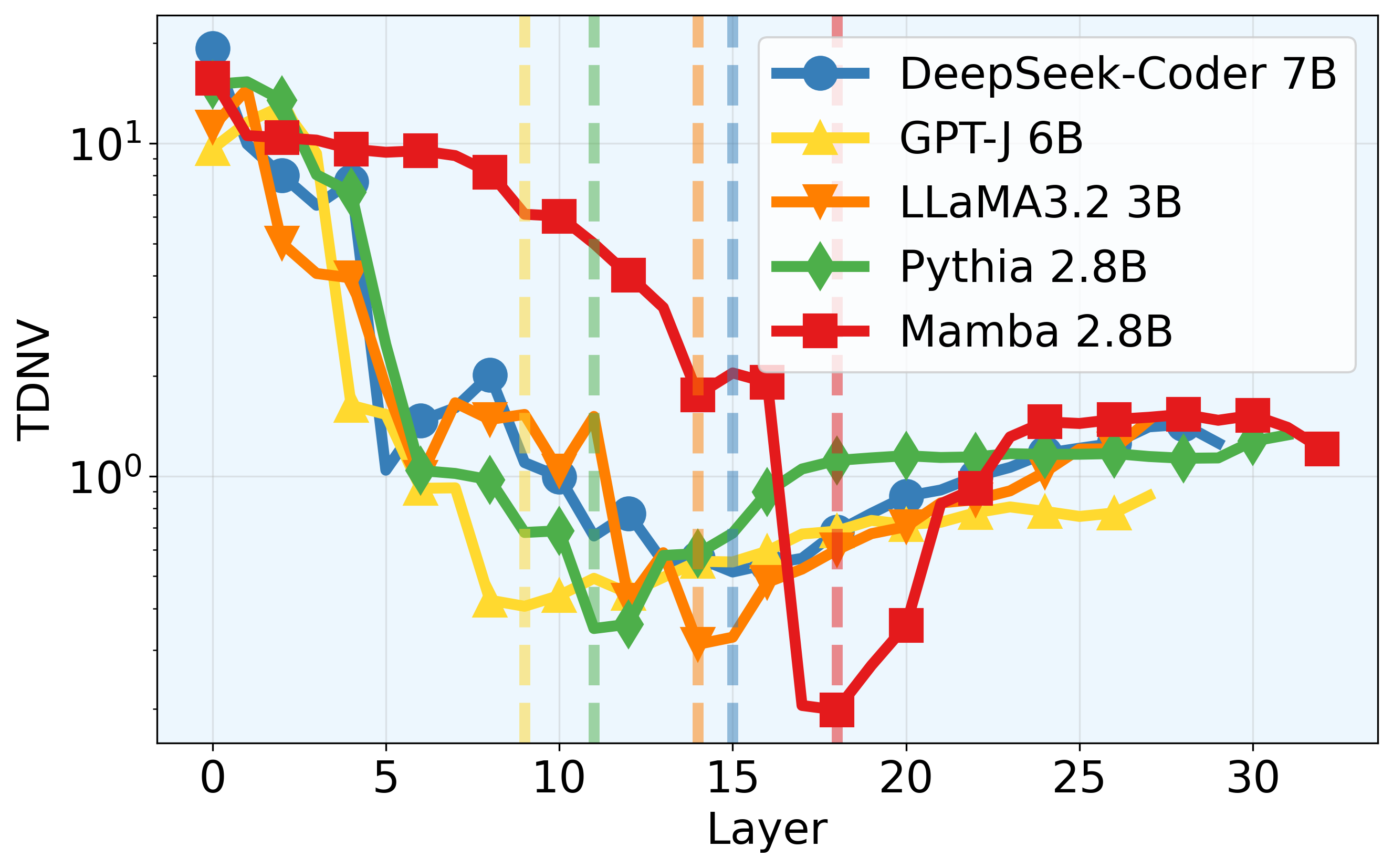}
        \caption{Layerwise TDNV of different model architectures, including decoder-only transformers and state-space models.}
        \label{fig:arch}
    \end{minipage}
    \hfill
    \begin{minipage}[t]{0.48\linewidth}
        \centering
        \includegraphics[height=4cm]{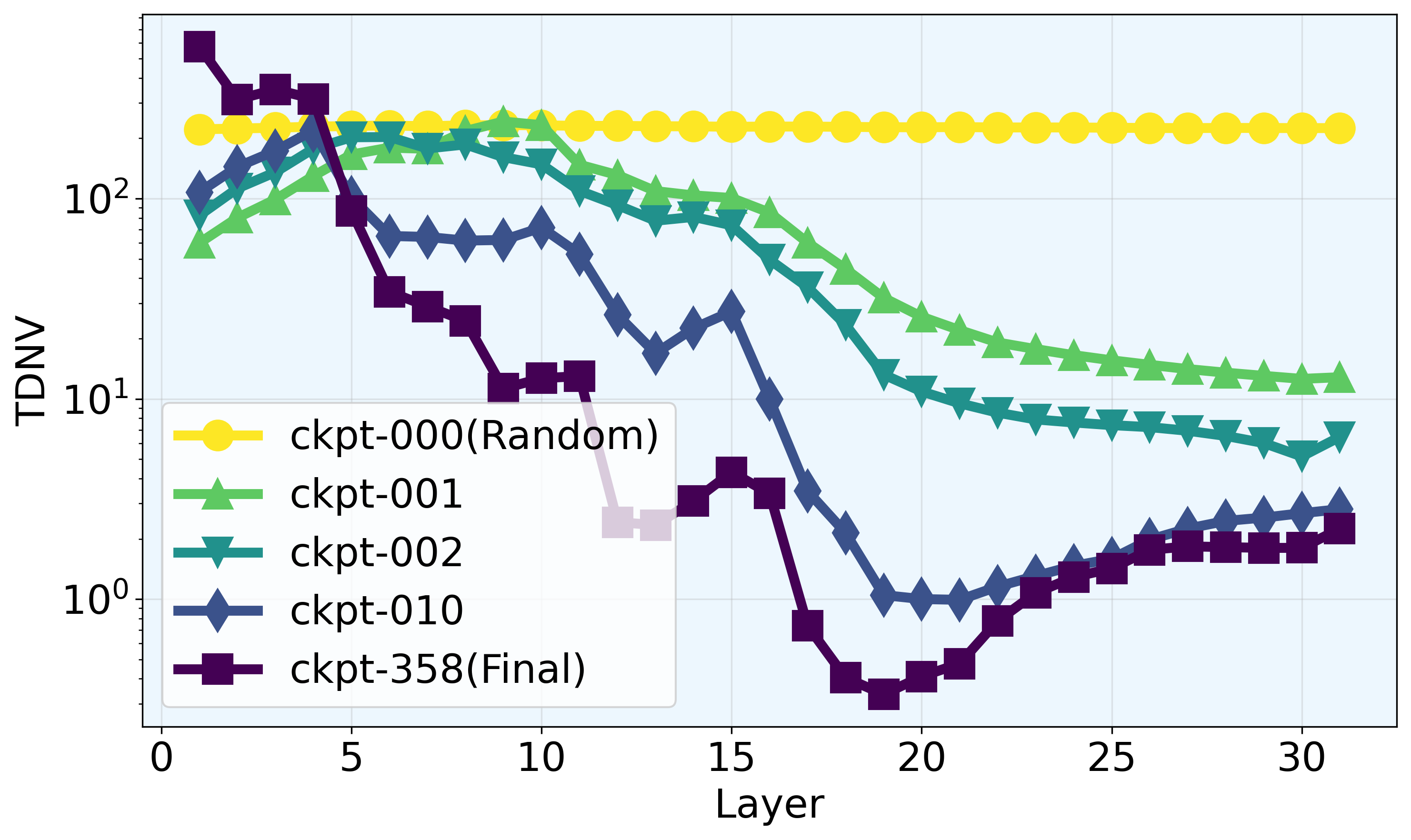}
        \caption{Layerwise TDNV during training process. The phenomenon emerges and intensifies with training progress. }
        \label{fig:tdnv-training}
    \end{minipage}
\end{figure}

\textbf{Universality across model architectures.} Following \citet{hendel2023context}, we first evaluate the algorithmic domain, including 5 tasks (copy letter, next letter, to uppercase, previous letter and next 2 letter). Detailed descriptions of all tasks are provided in \Cref{app:tasks}. As shown in \Cref{fig:arch}, the TDNV metric consistently exhibits a U-shaped trend—first decreasing then increasing—across two distinct architectural families: ($i$) Decoder-only transformers, including Llama3 \citep{grattafiori2024llama}, Pythia \citep{biderman2023pythia}, GPT-J \citep{gpt-j} and Deepseek-coder \citep{guo2024deepseek}. ($ii$) State-space models, specifically Mamba \citep{gu2023mamba}. This phenomenon holds even in the absence of attention, as evidenced by Mamba, indicating that the mechanism is not specific to the transformer architecture.

\begin{figure}[t]
    \centering

    \begin{minipage}[t]{0.32\linewidth}
        \centering
        \includegraphics[width=\linewidth]{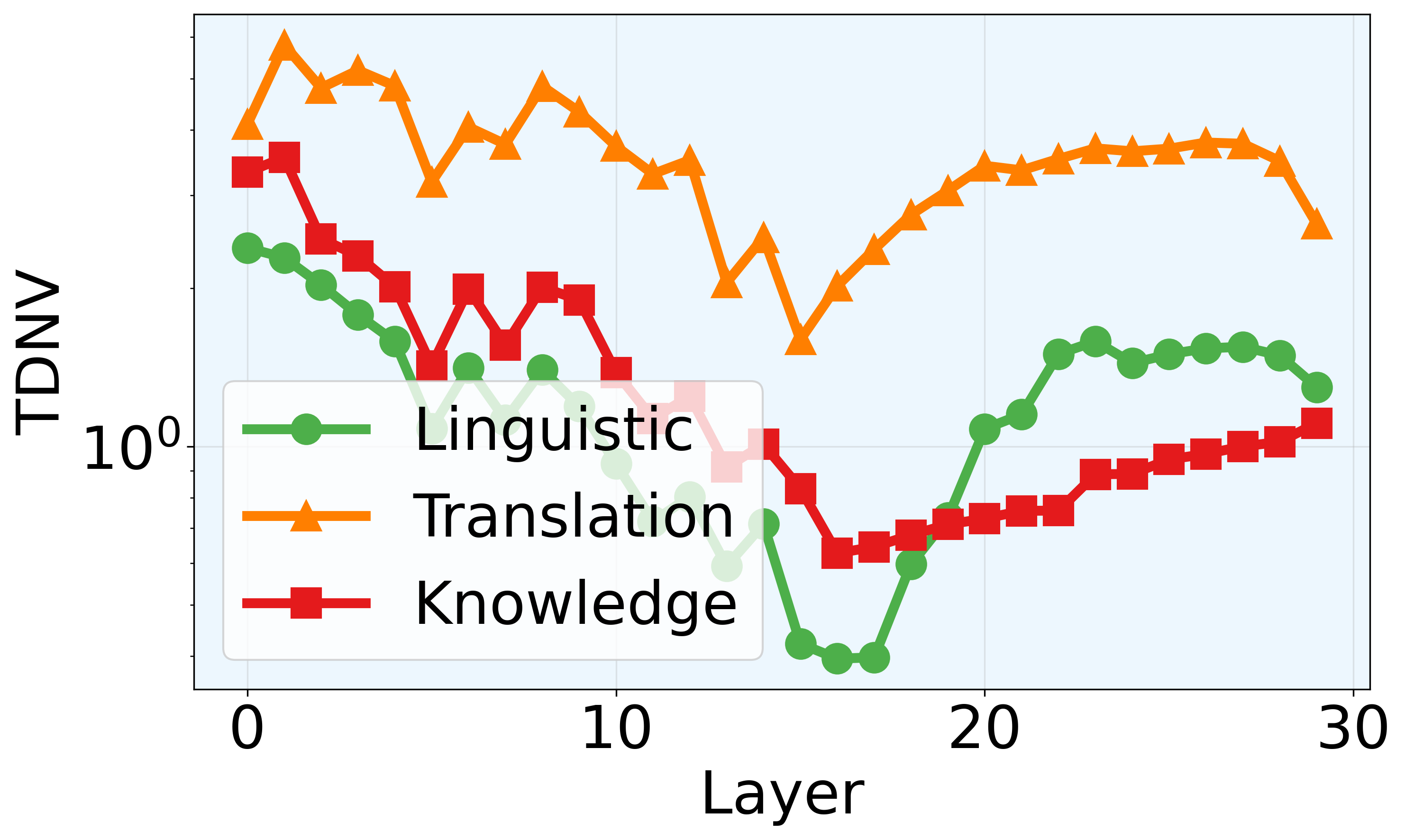}
        \caption{Symbolic ICL.}
        \label{fig:tdnv_symbolic}
    \end{minipage}
    \hfill
    \begin{minipage}[t]{0.32\linewidth}
        \centering
        \includegraphics[width=\linewidth]{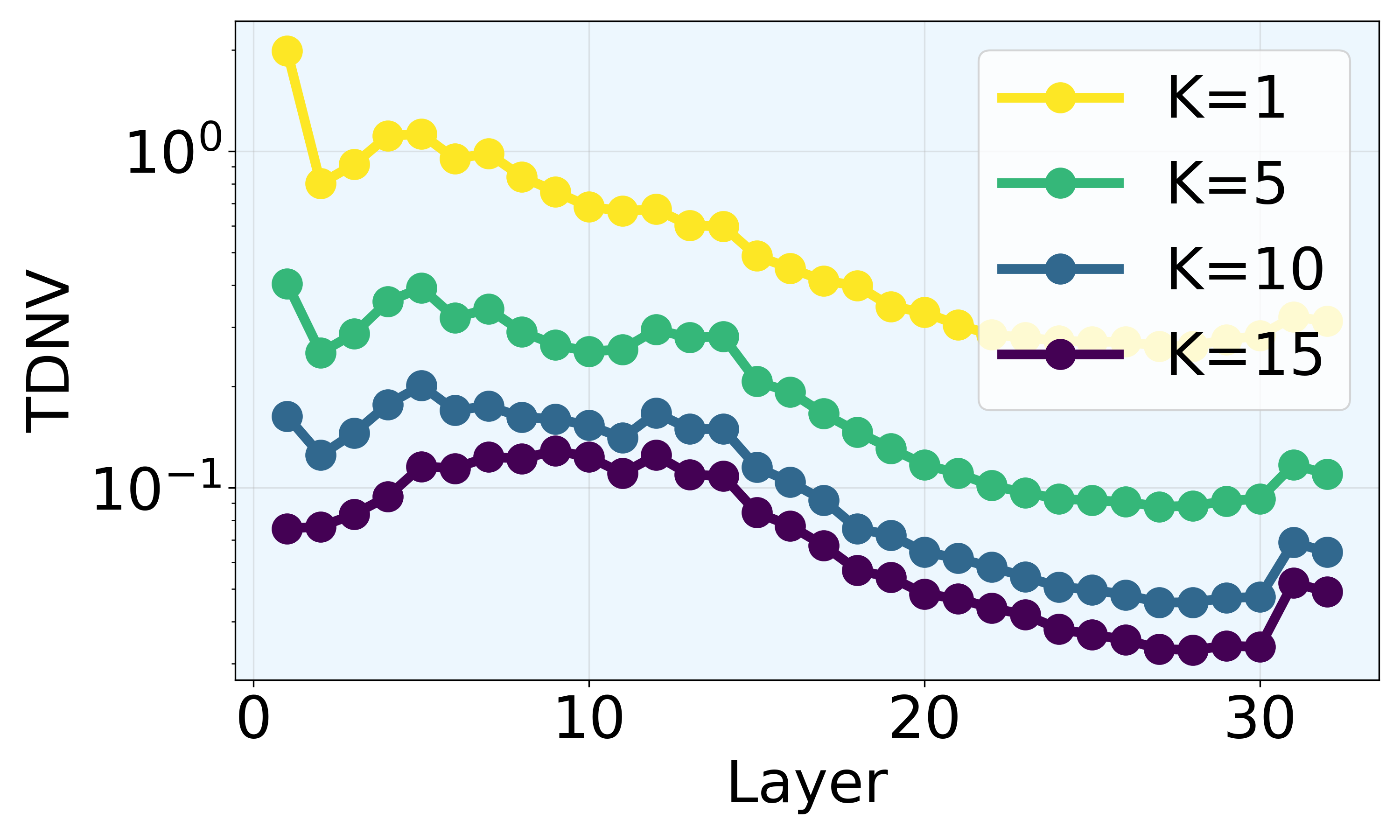}
        \caption{Language Understanding ICL.}
        \label{fig:tdnv_language}
    \end{minipage}
    \hfill
    \begin{minipage}[t]{0.32\linewidth}
        \centering
        \includegraphics[width=\linewidth]{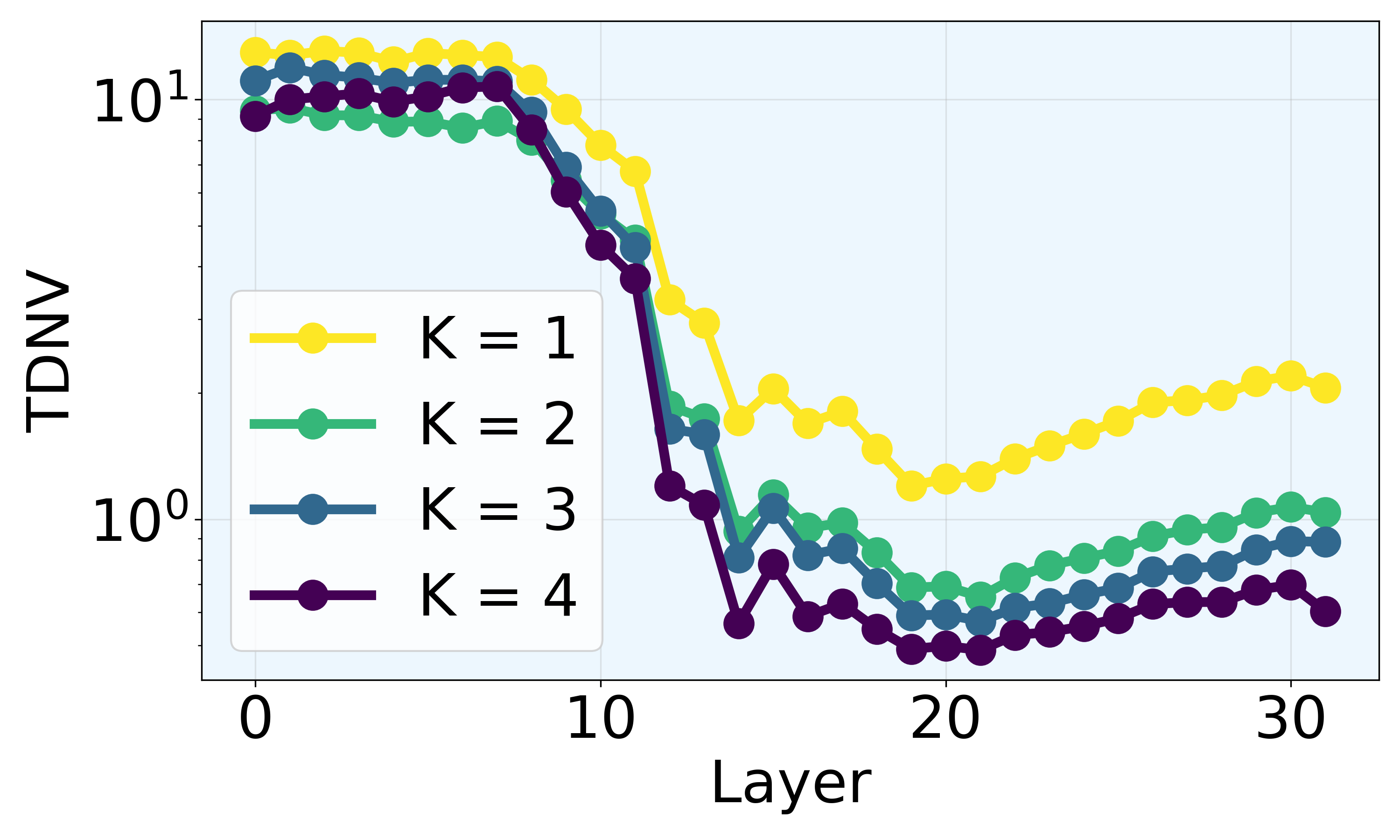}
        \caption{Multimodality ICL.}
        \label{fig:tdnv_MLLM}
    \end{minipage}
    
\end{figure}

\textbf{Universality across task domains.} To evaluate the generality of the phenomenon beyond algorithmic settings, we examine three additional task categories: ($i$) Symbolic ICL. We adopt the linguistic, translation, and knowledge domains from \citet{hendel2023context}, where TDNV consistently exhibits a U-shaped trend (\Cref{fig:tdnv_symbolic}). ($ii$) Language Understanding ICL. Beyond only short phrases, we evaluated TDNV on a natural language dataset with longer sentences. Each sentence can be analyzed across multiple attributes: length, semantic polarity, tense, sentence type, subject person, and entity type. We adopt Llama3 8B \citep{grattafiori2024llama} to predict the attribute label (e.g., positive or negative) for a query sentence based provided demonstrations with labels of a specific attribute (e.g., semantic polarity). As shown in \Cref{fig:tdnv_language}, the TDNV also exhibits a U-shaped trend, with the most compressed representation shifting to a later layer (around layer 28), indicating that longer sentences require more layers for effective task compression. ($iii$) Multimodality ICL. We further extend to a vision–language setting using a 2-D shape dataset, where each image contains a shape with four attributes (color, shape, size, texture). We adopt the Qwen-VL \citep{bai2023qwen} model to predict the attribute label (e.g., red or green) for a query image based provided demonstrations with labels of a specific attribute (e.g., color).  As shown in \Cref{fig:tdnv_MLLM}, the TDNV metric again exhibits a U-shaped curve. Across all settings, increasing the in-context length $K$ leads to more compact internal representations with lower TDNV values. Complete task specifications are provided in \Cref{app:tasks}.

\textbf{Emergence during training.} To verify that the phenomenon emerges only in trained models and not in randomly initialized ones, we evaluate on the Amber models of LLM360 family~\citep{liu2023llm360}, which provide intermediate checkpoints throughout the training process. As shown in \Cref{fig:tdnv-training}, models with random initialization exhibit flat TDNV values across all layers, indicating no structure of information compression. As training proceeds, the TDNV curve transitions into a distinct U-shape curve. This suggests the phenomenon only emerges as a result of training.

\subsection{Scaling Up Model size Leads to More Compression}
\label{subsec:model_size}
\begin{figure}[h]
  \centering
  \begin{minipage}[t]{0.49\linewidth}
    \centering
    \includegraphics[width=0.45\linewidth, trim=5 5 5 5, clip]{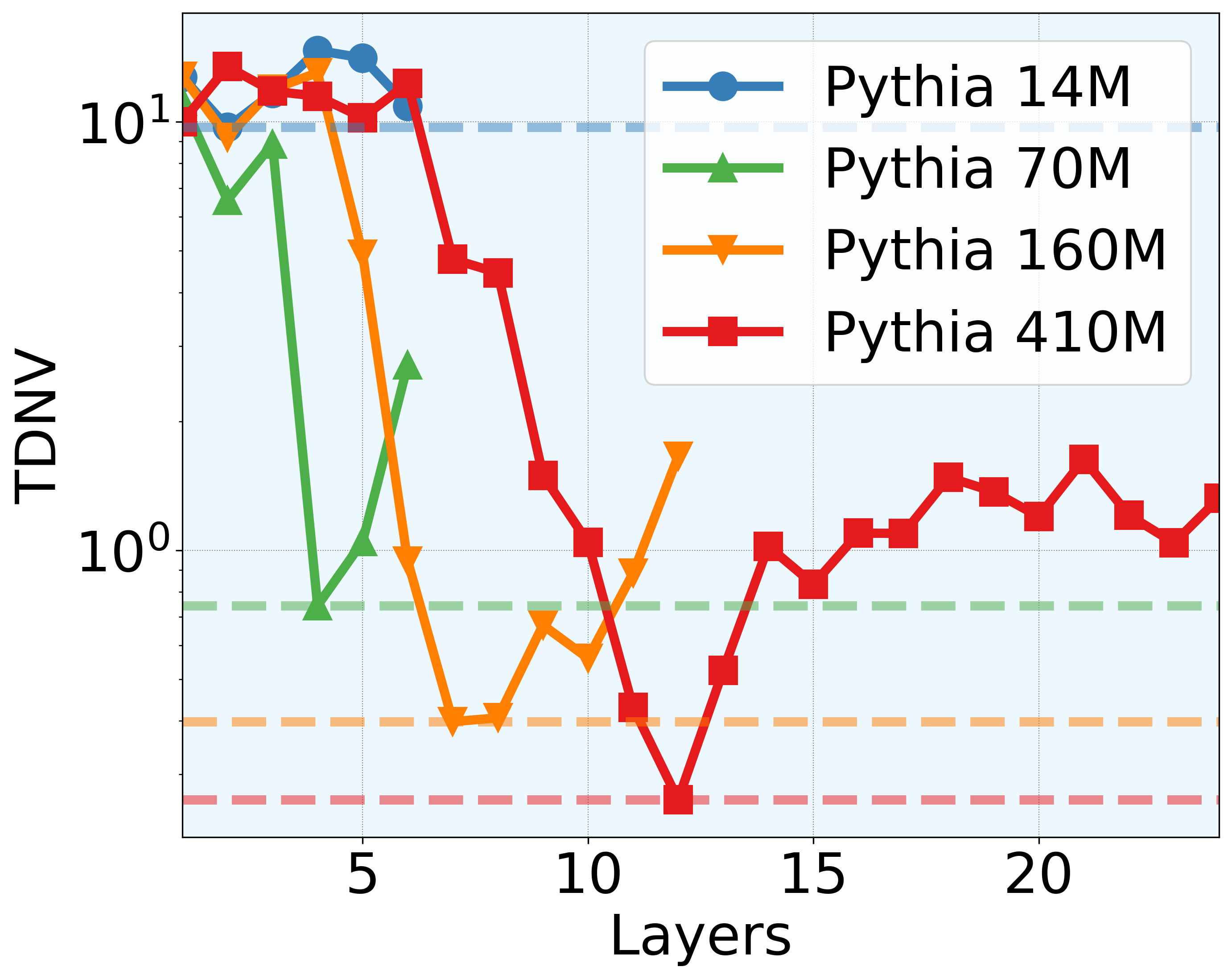}\hspace{0.5em}
    \includegraphics[width=0.48\linewidth, trim=5 5 5 5, clip]{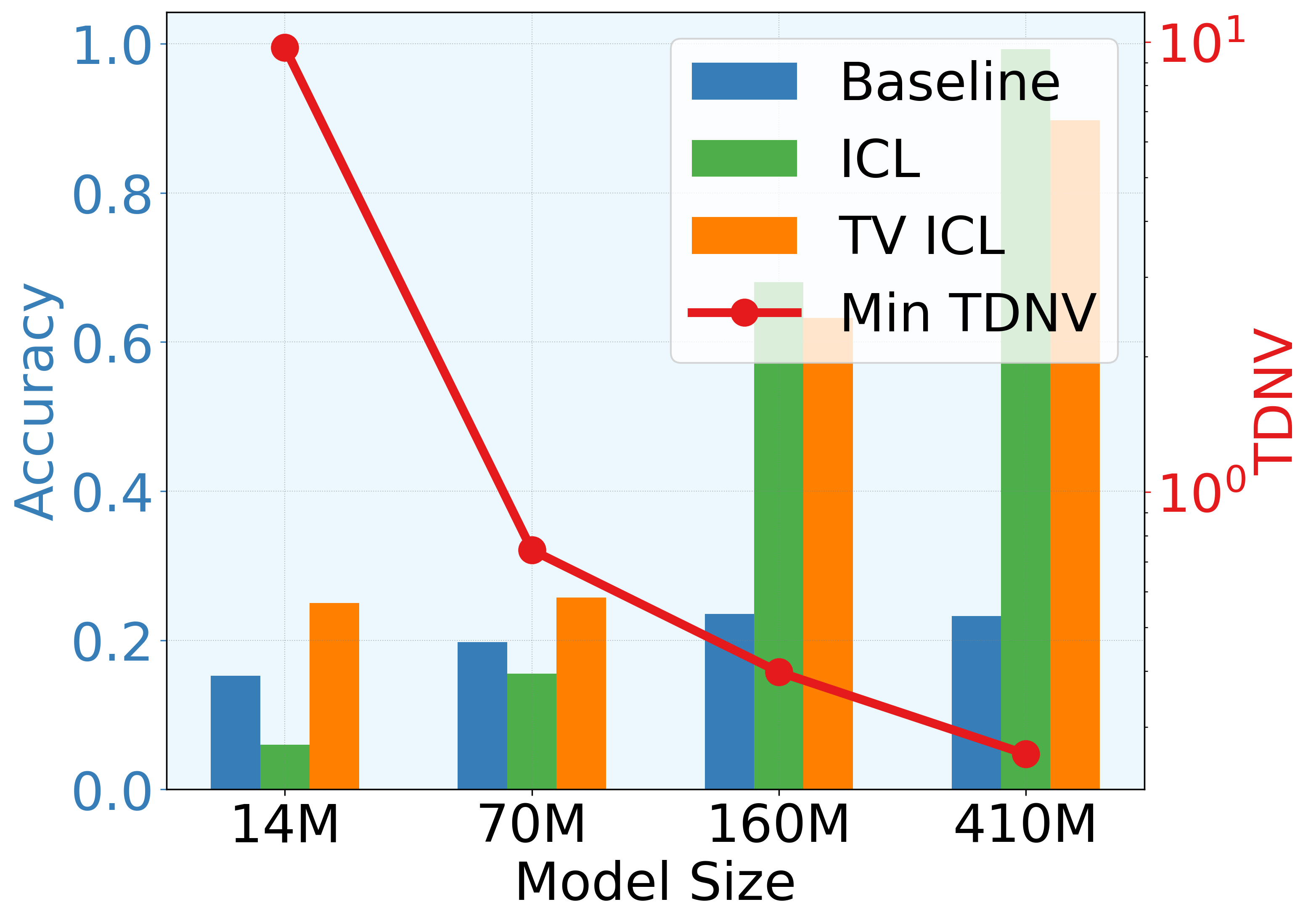}
    \caption{Effect of model size on layerwise TDNV and ICL performance.}
    \label{fig:model_size_group}
  \end{minipage}\hspace{0.5em}
  \begin{minipage}[t]{0.49\linewidth}
    \centering
    \includegraphics[width=0.48\linewidth, trim=5 5 5 5, clip]{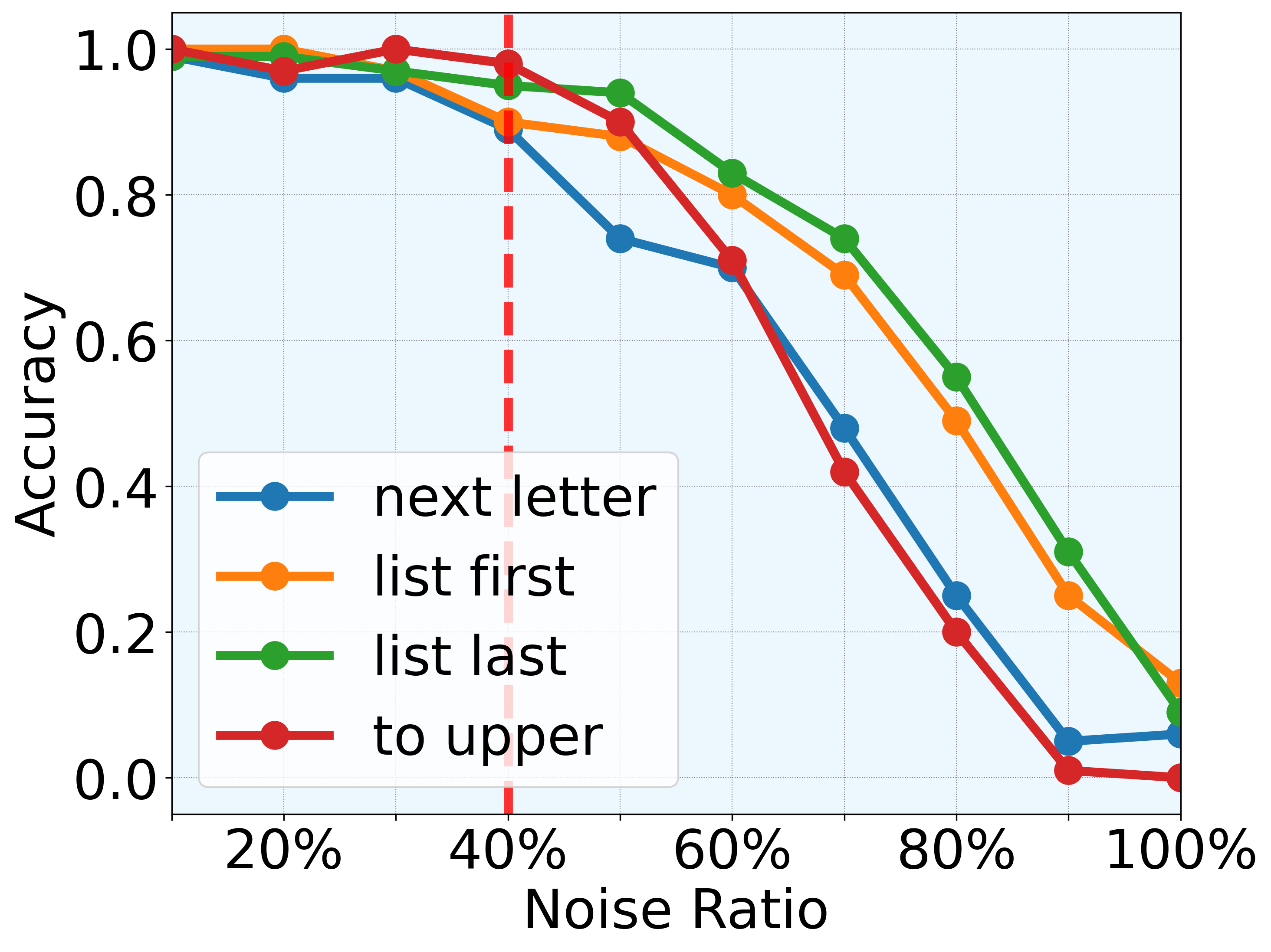}\hspace{0.5em}
    \includegraphics[width=0.48\linewidth, trim=5 5 5 5, clip]{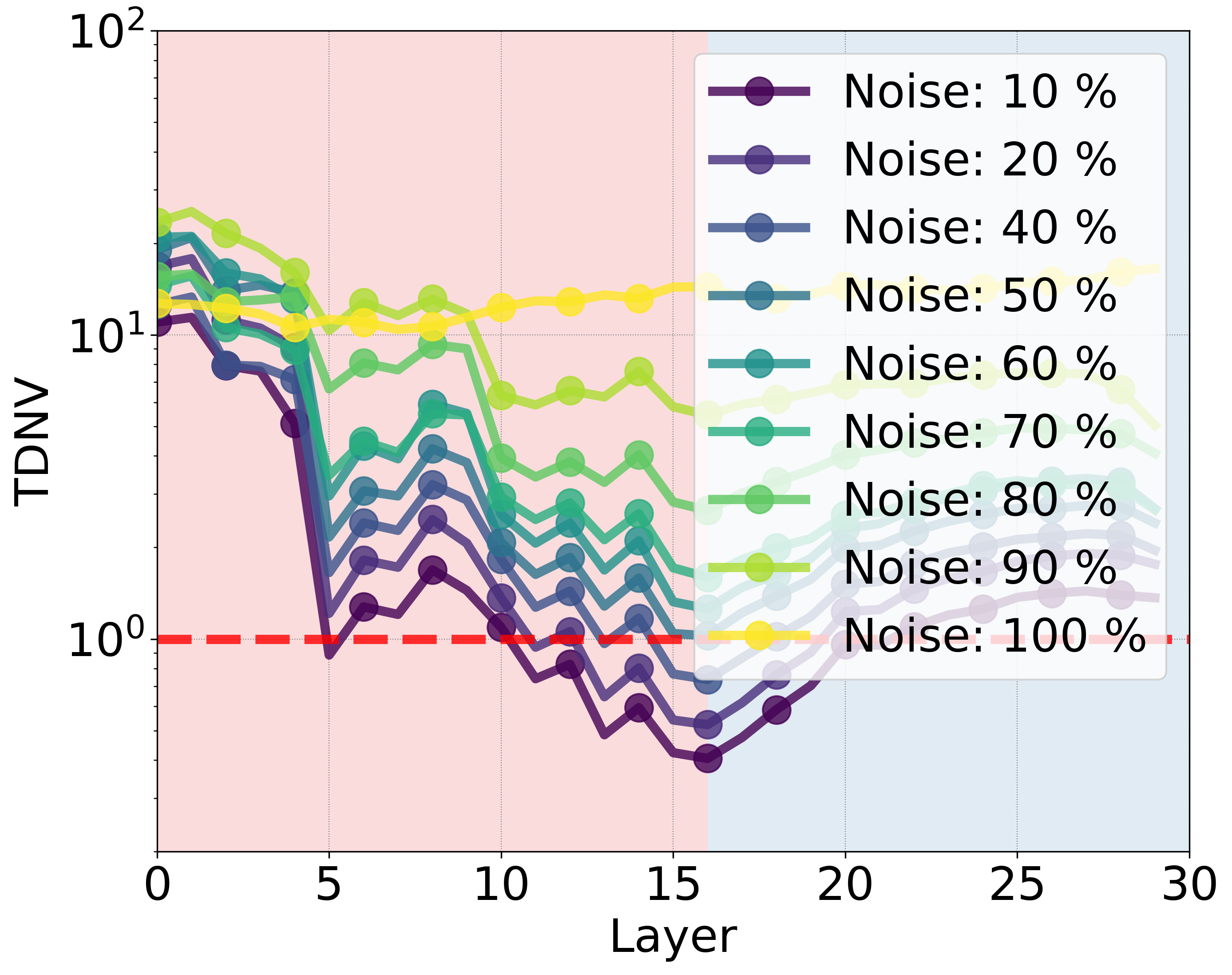}
    \caption{Effect of noisy demonstrations on ICL performance and layerwise TDNV.}
    \label{fig:noise_group}
  \end{minipage}
\end{figure}

To explore how model size influences information compression, we analyze Pythia models ranging from 14M to 410M parameters in terms of both layerwise TDNV and performance (as shown in \Cref{fig:model_size_group}). We evaluate ICL performance from two perspectives: (1) the regular few-shot setting, referred to as ICL, and (2) the task-vector (TV) setting---i.e., zero-shot ICL using a task vector patched from the best-performing layer $\hat{\ell}$ identified under the few-shot setting---referred to as TV ICL. Higher accuracy in either setting indicates better performance. Additionally, we report zero-shot accuracy without any task-vector information, referred to as the baseline. We find that larger models tend to produce more compressed and discriminative representations in the middle layers, indicating a stronger ability to extract task information from the demonstrations, thereby achieving better performance in both ICL and task-vector ICL.

\subsection{Compression-to-Expression under Noisy Demonstrations}
\label{sec:robustness}

Demonstrations in real-world scenarios are often noisy, with some input-output pairs failing to accurately represent the intended task. Despite this, ICL demonstrates notable robustness to such noisy demonstrations. As illustrated in \Cref{fig:noise_group} left, performance remains largely unaffected even when the noise ratio reaches up to 40\%, where the noise ratio is defined as the proportion of incorrect input-output pairs relative to the total number of pairs. To understand this robustness, we explore it through the lens of information compression.

We plot the layerwise TDNV under varying noise ratios in \Cref{fig:noise_group} right and highlight two key observations: ($i$) higher noise ratios consistently lead to increased TDNV across all layers, indicating that noisy demonstrations impair the model's ability to compress and extract task-relevant information. In the extreme case of 100\% noise—where inputs and labels are completely uncorrelated—the model receives no meaningful task signal, and the characteristic compression-to-expression pattern disappears across layers. ($ii$) When the noise ratio remains below 40\%, the minimum TDNV values stay below 1, indicating that within-task variance is still smaller than between-task distance. This allows task representations to remain distinguishable, resulting in minimal performance degradation. This observation explains the robust performance at noise ratios below 40\%. However, beyond 40\% noise, task representations become increasingly entangled, causing performance to decline rapidly.

\paragraph{The position of noisy demonstration affects compression.}

\begin{figure}[t]
    \centering
    \begin{minipage}[t]{0.33\linewidth}
        \centering
        \includegraphics[width=\linewidth]{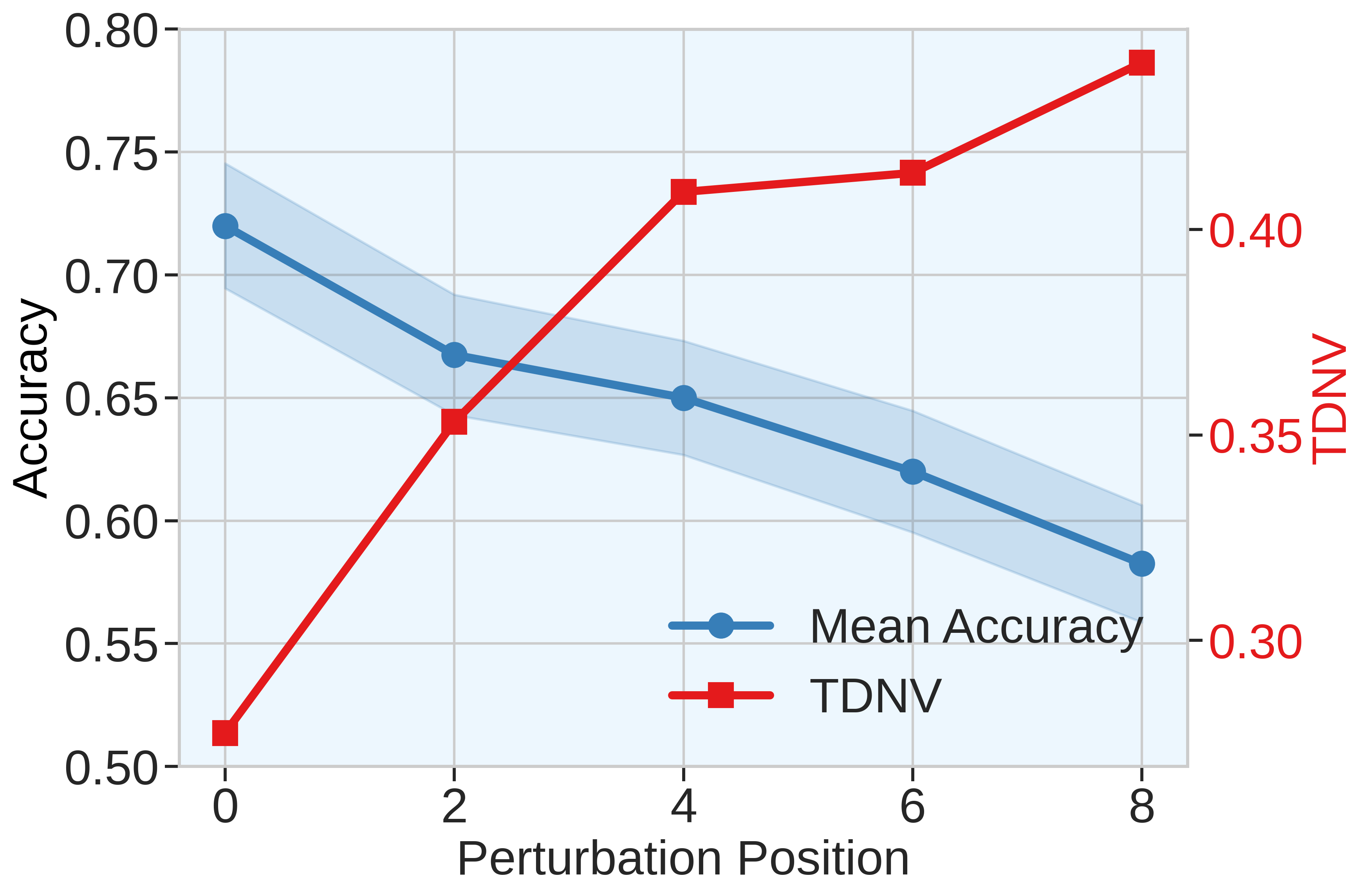}
        \caption{Layerwise TDNV and ICL accuracy under different perturbation positions.}
        \label{fig:acc_tdnv}
    \end{minipage}
    \hfill
    \begin{minipage}[t]{0.33\linewidth}
        \centering
        \includegraphics[width=\linewidth]{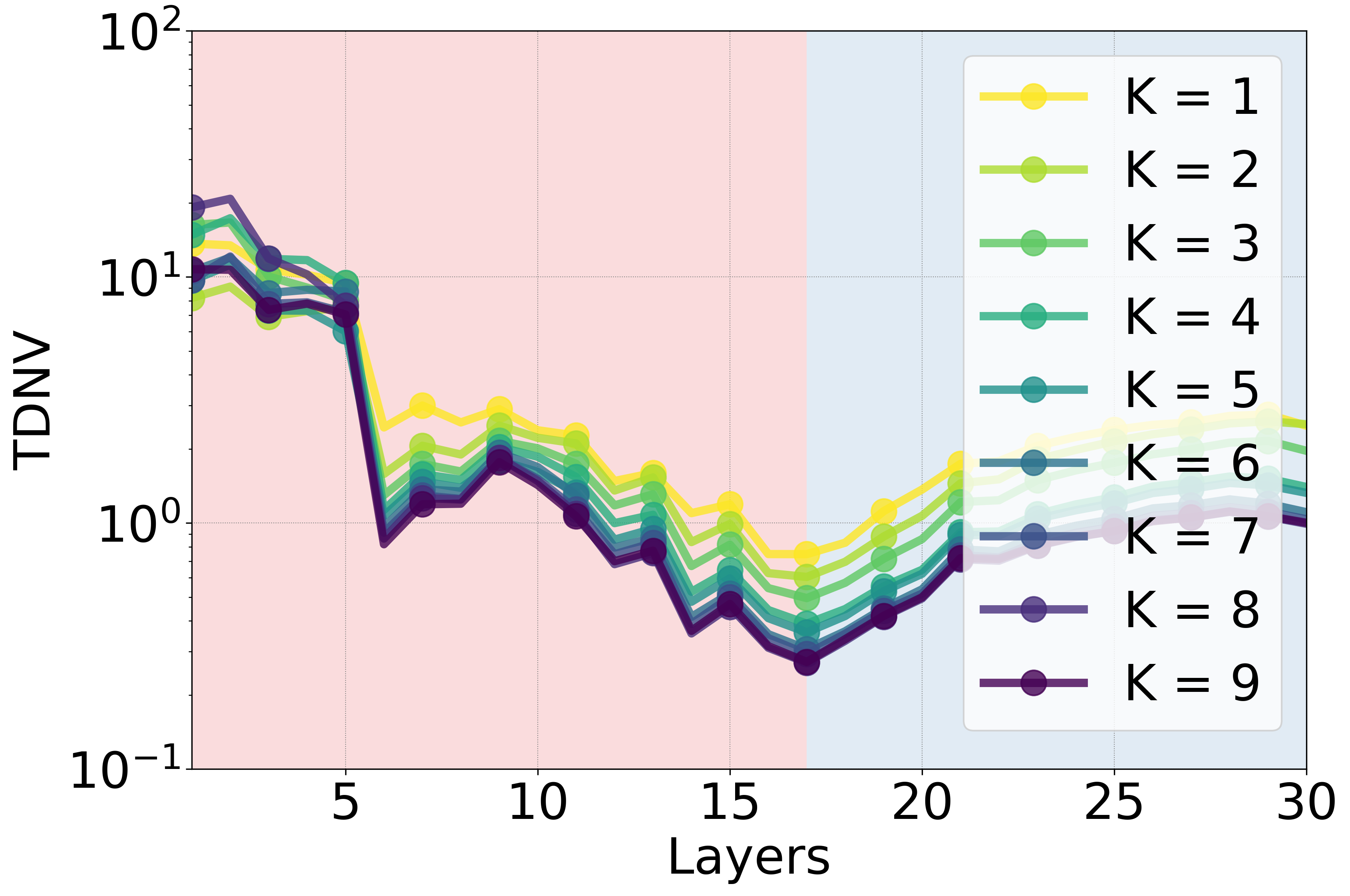}
        \caption{Layerwise TDNV under different number of demonstrations $K$.}
        \label{fig:num_demon}
    \end{minipage}
    \hfill
    \begin{minipage}[t]{0.30\linewidth}
        \centering
        \includegraphics[width=\linewidth]{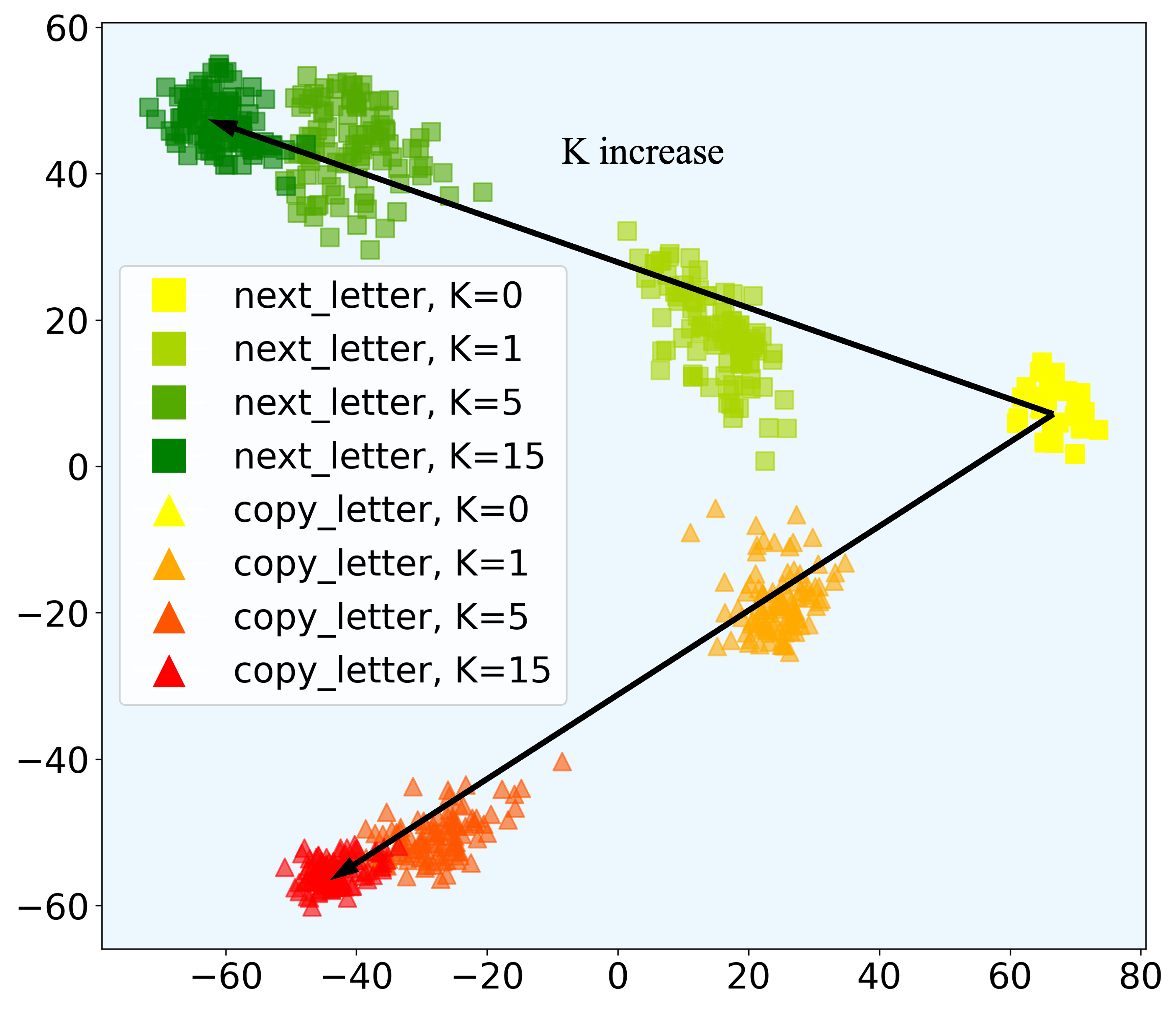}
        \caption{PCA of task vectors as number of demonstrations $K$ increase.}
        \label{fig:pca}
    \end{minipage}
\end{figure}

Unlike conventional machine learning, the order of demonstrations has a significant impact on model performance in ICL \citep{liu2023lost, lu2021fantastically, zhou2024detail}. As illustrated in \Cref{fig:acc_tdnv}, perturbing demonstrations that appear later in the sequence causes a larger performance drop and higher TDNV values. These perturbations result in less compressed task vectors, indicating that later demonstrations play a more crucial role than earlier ones in helping the model extract task information. To gain deeper insight, we perform a fine-grained analysis by computing layerwise TDNV for each separator token, referred to as grid TDNV in \Cref{app:grid_tdnv}.

\section{Bias-variance Decomposition of Task Vectors}
\label{subsec:bias-var}

In this section, we analyze the effect of the number of demonstrations and study the task vectors in the middle layer---where the representation is most compact (i.e., exhibits the smallest TDNV)---using a bias–variance decomposition.

\paragraph{Increasing in-context lengths lead to more compressed representations.} We first evaluate how the number of demonstrations affects the geometry of ICL representations using layerwise TDNV. The results in \Cref{fig:num_demon} show that increasing the number of demonstrations $K$ consistently reduces TDNV across all layers. This indicates that as more demonstrations are provided, the within-task variance of task vectors decreases while the between-task distance increases. This explains why increasing the number of demonstrations improves performance---it leads to more compressed and more distinct representations in the intermediate layers.

\textbf{Bias-Variance decomposition.} 
In \Cref{fig:pca}, we present a PCA visualization of the most compressed layer for two different tasks that share the same query; see an illustrative example in \eqref{eq:example-tasks}. When no demonstrations are provided ($K = 0$), both tasks produce the same vectors that reflect the prior of the pretrained model. As $K$ increases, we observe an intriguing phenomenon:
$(i)$ different tasks induce task vectors in distinct directions, yet each task follows a consistent direction; $(ii)$ the variance within each task decreases. Based on this observation, we decompose the task vector $\vh_{i,t}(K)$ (where we highlight the dependence on the number of demonstrations $K$ and omit the superscript $(\ell)$) of each instance into the following components
\begin{equation}
    \vh_{i,t}(K)= \boldsymbol{\mu}_{t}(\infty) + \underbrace{\boldsymbol{\mu}_t(K) - \boldsymbol{\mu}_{t}(\infty)}_{\text{bias}} +  \underbrace{\vh_{i,t}(K) - \vmu_{t}(K)}_{\text{variance}},
\end{equation}
where $\vmu_t(K) = \E_i [\vh_{i,t}(K)]$ denotes the mean of the task vector obtained from $K$ demonstrations, and $\vmu_t(\infty) = \lim_{K\rightarrow \infty}\E_i [\vh_{i,t}(K)]$ represents the mean of the task vector obtained from infinitely many possible demonstrations (ignoring the practical limitations of context length in real-world LLMs), which maybe referred to as the {\it ideal} task vector.

\begin{figure}[t]
    \centering

    \begin{minipage}[t]{0.32\linewidth}
        \centering
        \includegraphics[width=\linewidth]{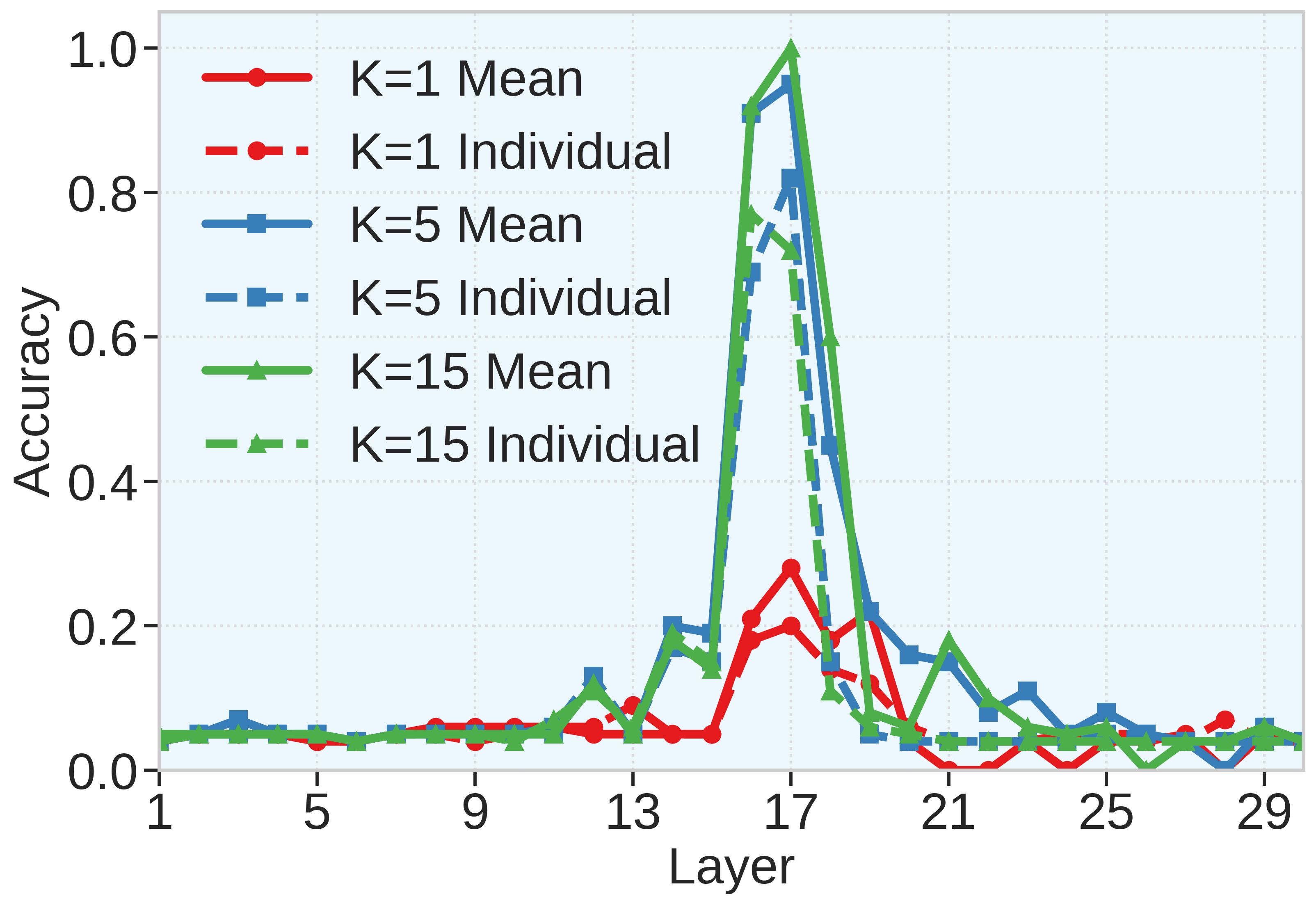}
        \caption{Layerwise task vector accuracy using the mean v.s. individual task vectors.}
        \label{fig:inject_mean}
    \end{minipage}
    \hfill
    \begin{minipage}[t]{0.32\linewidth}
        \centering
        \includegraphics[width=\linewidth]{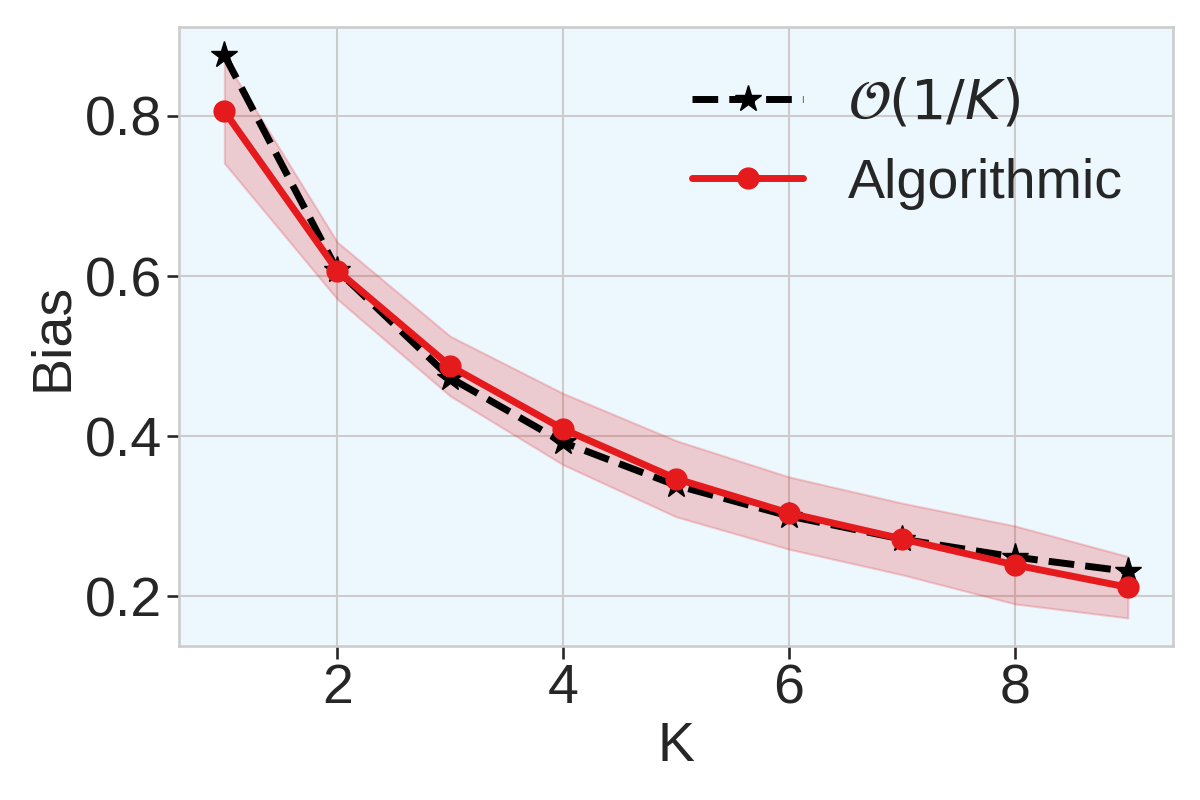}
        \caption{Decrease of bias at rate of $\mc O(1/K)$.}
        \label{fig:distance_decay}
    \end{minipage}
    \hfill
    \begin{minipage}[t]{0.32\linewidth}
        \centering
        \includegraphics[width=\linewidth]{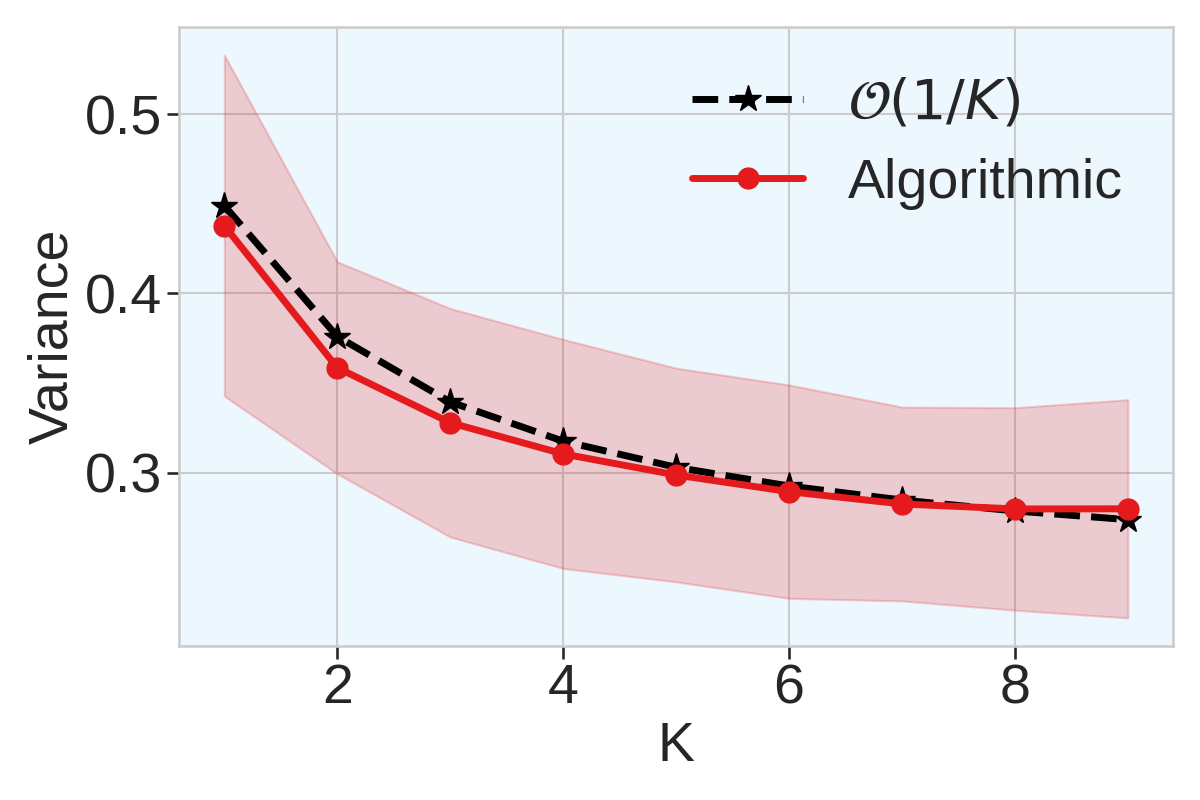}
        \caption{Decrease of variance at rate of $\mc O(1/K)$.}
        \label{fig:var_decay}
    \end{minipage}
\end{figure}

\paragraph{How well does the mean task vector encode task information?}
Unlike the classical bias–variance decomposition---where the mean of multiple models often outperforms individual models due to the ensemble effect---the setting here is more nuanced. The mean task vector $\boldsymbol{\mu}(K)$ is averaged not only over different demonstrations but also over different queries. Therefore, it is not immediately clear whether the mean task vector still encodes useful task information---and if it does, whether it does more effectively than individual vectors \( \vh_i(K) \). To address this question, we compare task vector accuracy \citep{hendel2023context} using both individual task vectors and the mean task vector, with results shown in \Cref{fig:inject_mean}. Remarkably, we observe that injecting the mean task vector consistently leads to better performance, suggesting that it encodes task information more effectively than individual task vectors. Moreover, the performance of the mean task vector also exhibits an inverted U-shaped curve: it peaks around the most compressed layer and improves as the number of demonstrations increases. Together with the results in \Cref{fig:num_demon,fig:pca}, this indicates that---ignoring the practical limitations of context length---ICL representations constructed from infinitely many possible demonstrations also exhibit the \phenomenon phenomenon, and the corresponding task vector $\vmu(\infty)$ at the most compressed layer can thus be viewed as an ideal task representation. This suggests a promising direction for a theoretical investigation of the phenomenon in the infinite-sample regime, which we leave for future work. 

Building on this understanding, we now use the bias–variance decomposition to study how the number of demonstrations $K$ influences task vectors at the most compressed layer.

\begin{itemize}
[leftmargin=12pt,itemsep=2pt,topsep=0pt,parsep=0pt]
    \item {\bf Decrease of bias:} The task mean vector $\vmu_t(K)$ progressively shifts from the zero-shot mean vector $\vmu_t(0)$ (encodes the pertaining bias) toward the ideal task vector $\vmu_t(\infty)$ as $K$ increases, capturing more information about the task. In other words, the bias term $\vmu_t(K) - \vmu_t(\infty)$ decreases with increasing $K$, and empirically, as shown in \Cref{fig:distance_decay}, we observe that $\|\vmu_t(K) - \vmu_t(\infty)\|_2/ \|\vmu_t(0) - \boldsymbol{\mu}_t(\infty)\|_2$ decays roughly at a rate of $\mc O(1/K)$.
    \item {\bf Decrease of variance:}  On the other hand, as shown in \Cref{fig:var_decay},  the variance term $\|\vh_{i,t}(K) - \vmu_{t}(K)\|_2^2$ decays roughly at a rate of $\mc O(1/K)$, which together with the fact that between-task distance becomes a constant when $K$ is large leads to a decay rate of $\mc O(1/K)$ for TDNV.
\end{itemize}
We remark that, unlike the classical bias–variance decomposition of prediction risk in terms of model capacity---where a trade-off may exist---our bias–variance decomposition applies to the task vector in ICL and exhibits no such trade-off with respect to the number of demonstrations $K$. In other words, both the bias and variance decrease as $K$ increases, indicating that the task vector converges to an ideal task representation. This further suggests that LLMs generate increasingly compact and informative representations from the input demonstrations in the compression phase, with the compact representation encoding task information and converging as the number of demonstrations becomes sufficiently large.

\textbf{Theoretical analysis of bias-variance terms for attention layer.} To develop a theoretical understanding of how the bias and variance terms of task vectors evolve with the number of demonstrations $K$, we consider a simplified setting.
Specifically, we analyze a single-layer attention model, as the attention mechanism plays a central role in extracting task representations from demonstrations. To simplify the presentation, we drop the subscript $t$ as we only focus on one task and assume that each demonstration and the query correspond to a single hidden state, denoted as \( \vh_1, \ldots, \vh_K \in \R^d \) for the demonstrations and \( \vh_q \in \R^d \) for the query, respectively. To facilitate analysis, we adopt a linear attention mechanism, denoted by $\mathrm{Attn}$, which preserves the normalization property of softmax attention, namely, that the attention weights sum to 1 \citep{katharopoulos2020transformers, shen2021efficient}. Linear attention has been widely adopted in the literature for theoretical analysis of ICL~\citep{von2023transformers,ahn2023transformers,wang2024transformers,li2024context}.
\begin{theorem}[Bias-variance decomposition with respect to $K$] Suppose that each demonstration $\vh_i,i =1,\ldots,K$ is i.i.d. randomly generated from a distribution $\calH$ on $\R^d$. Then the output of the query token, $
    \vh'_{q}(K) = [\mathrm{Attn}\left(\vh_1,\ldots,\vh_K, \vh_{q})\right)]_{:,K+1}$,
where $[]_{:,K+1}$ means the $(K+1)$-th column, satisfies the following statistical properties as the number of demonstration $K$ increases:
\begin{itemize}
[leftmargin=12pt,itemsep=2pt,topsep=0pt,parsep=0pt]
    \item \textbf{(Decrease of variance)} The variance decays as $
        \mathrm{Var}(\|\vh'_{q}(K)\|_2^2) = \mathcal{O}(1/K)$.
    
    \item \textbf{(Decrease of bias)} The mean output $\mathbb{E}[\vh_{q}'(K)]$ evolves as a linear combination of the zero-shot mean $\mathbb{E}[\vh_{q}'(0)]$ and infinite-shot mean $\mathbb{E}[\vh_{q}'(\infty)]$:
    \begin{equation}
        \mathbb{E}[\vh_{q}'(K)] = \lambda_K\, \mathbb{E}[\vh_{q}'(0)] + (1 - \lambda_K)\, \mathbb{E}[\vh_{q}'(\infty)],
    \end{equation}
   which further implies that the bias term decays as $\|\mathbb{E}[\vh_{q}'(K)] - \mathbb{E}[\vh_{q}'(\infty)]\|_2 = \mathcal{O}(1/K)$.
\end{itemize}
\label{theorem:bias-var}
\end{theorem}
The proof for \Cref{theorem:bias-var} can be found in \Cref{app:proof}. While our analysis shares the same simplifications and limitations as prior work on linear self-attention ~\citep{von2023transformers,ahn2023transformers,wang2024transformers,li2024context}, it offers new insights into the functional role of attention in ICL, revealing how attention contribute to reducing both the variance and bias of task representations---leading to improved performance as the number of demonstrations increases.

\begin{figure}[t]
    \centering

    \begin{minipage}[t]{0.4\linewidth}
        \centering
        \includegraphics[width=\linewidth]{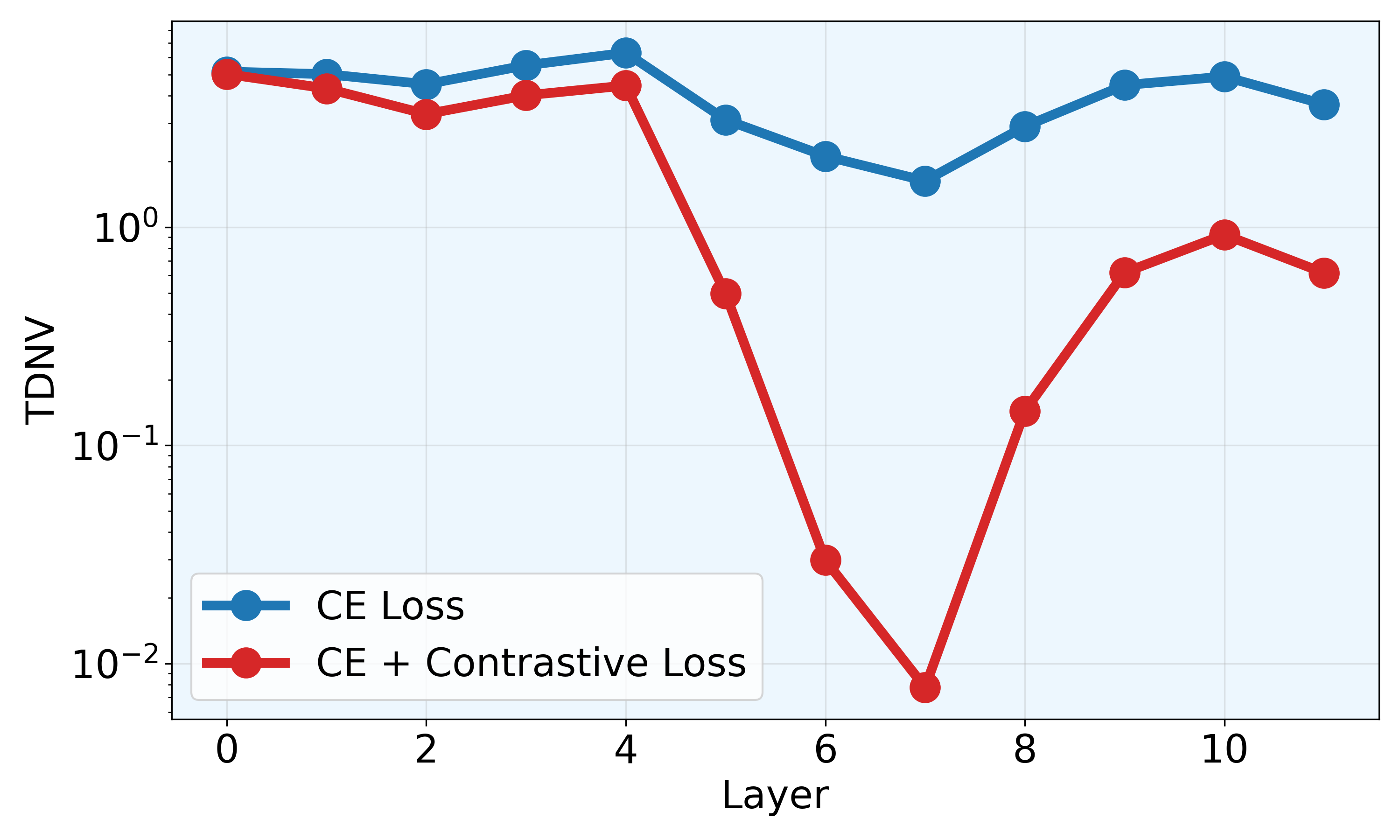}
        \caption{Layerwise TDNV using model trained with CE loss v.s. CE + contrastive loss on layer $7$.}
        \label{fig:tndv_gpt2}
    \end{minipage}
    \hfill
    \begin{minipage}[t]{0.4\linewidth}
        \centering
        \includegraphics[width=\linewidth]{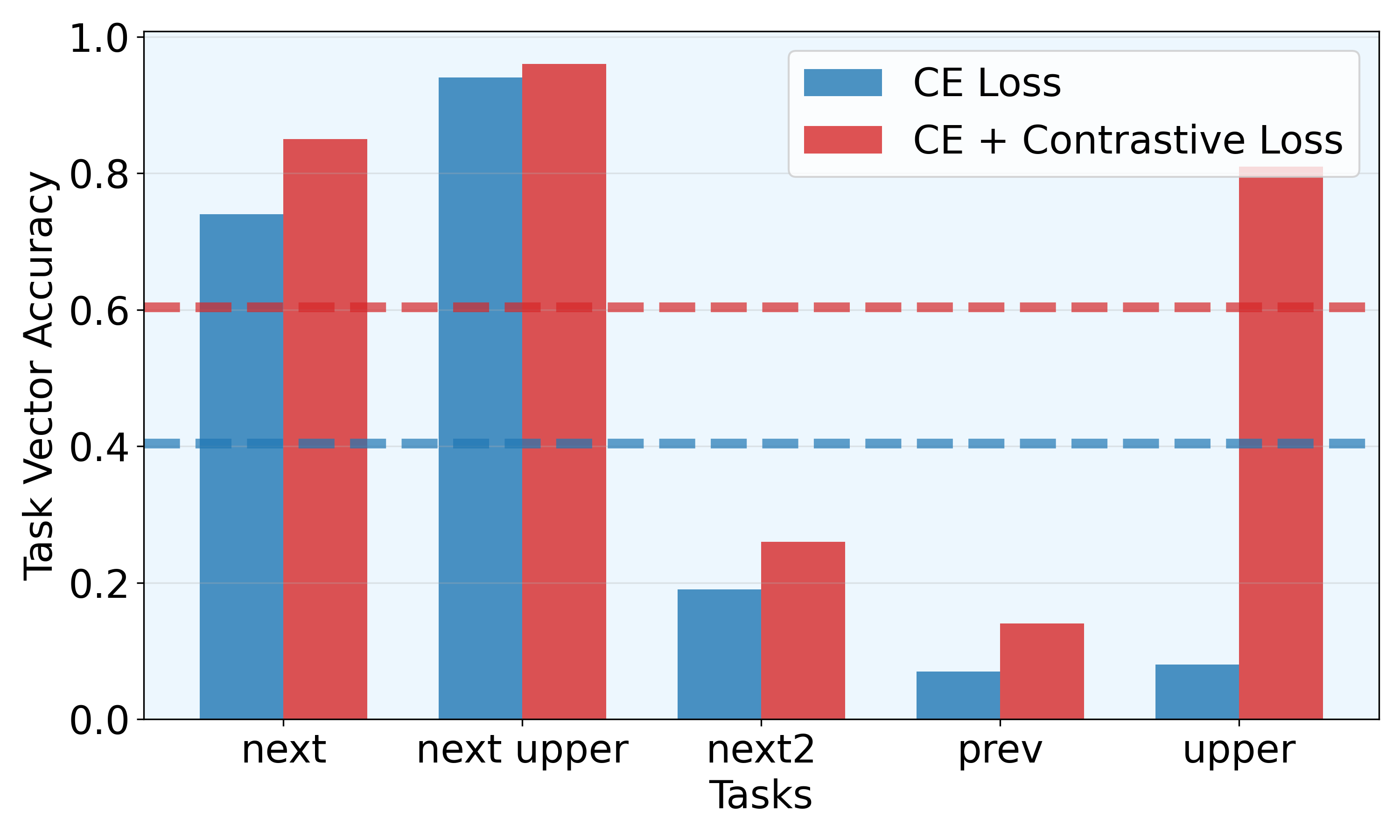}
        \caption{Task-vector contrastive fine-tuning improves task-vector accuracy.}
        \label{fig:tv_acc_gpt2}
    \end{minipage}
\end{figure}

\section{Applications of Compression-to-Expression}

\textbf{Identify the optimal task vector layer.} To find the optimal intermediate layer for task vector extraction, previous works \citep{hendel2023context} typically patch the vector at each layer and evaluate accuracy using a validation set. TDNV provides a more efficient way to identify the optimal layer for task vectors. As shown in \Cref{fig:main}, at a certain intermediate layer $\hat{\ell}$, we observe three simultaneous changes: the TDNV shifts from decreasing to increasing, the task vector accuracy begins to decrease, and the early-exit accuracy starts to increase. Since the layer with minimum TDNV corresponds to the layer with maximum task vector accuracy, we can identify the optimal layer with just one pass of inference using $
\hat{\ell} = \argmin_{\ell \in [1, L]} \text{TDNV}^{(\ell)}$.

\textbf{Task-vector contrastive fine-tuning improves task-vector accuracy.} Motivated by prior evidence that more compressed task vectors yield better performance, we propose task-vector contrastive fine-tuning that explicitly encourages such compression. Specifically, during fine-tuning on ICL tasks, we augment the cross-entropy (CE) loss with a contrastive loss applied to the task vectors. This loss pulls representations from the same task closer together while pushing apart those from different tasks (see \Cref{app:enhance_tv} for the exact formulation). We fine-tune a pretrained GPT-2 model on symbolic ICL domains using either the baseline CE loss or our combined loss applied in $7$-th layer. As shown in \Cref{fig:tndv_gpt2}, the contrastive term lowers TDNV, indicating stronger task-vector compression, which in turn boosts downstream task-vector accuracy by an average of 20\% (\Cref{fig:tv_acc_gpt2}).

\section{Conclusion}
\label{sec:conclusion}

This work provides a comprehensive analysis of the internal dynamics of ICL in LLMs. We uncover a prevalent \textit{\phenomenon} phenomenon in ICL representations, shedding light on how task information is compressed and later expressed to generate predictions. We show that it has profound implications for ICL performance and robustness and reveal the role of attention mechanisms. These insights not only deepen our understanding of structured representation learning in LLMs but also offer practical implications for improving interpretability, efficiency, and robustness.

\section{Acknowledgement}
We acknowledge support from NSF grants IIS-2312840 and IIS-2402952, as well
as the ORAU Ralph E. Powe Junior Faculty Enhancement Award. We gratefully acknowledge Xia Ning, Qing Qu, Peng Wang, and Xifeng Yan for valuable discussions.

\bibliographystyle{unsrtnat}
\bibliography{reference}

\appendix

\newpage

\begin{center}
    {\LARGE \bfseries Appendix}
\end{center}

The appendix is organized as follows. We first provide detailed descriptions of all tasks in \Cref{app:tasks}. Next, we add more discussion on related works in \Cref{app:related_works}. \Cref{app:additional_exp} presents additional experiments, including verification of the \textit{\phenomenon} and an ablation study on the choice of the last separation tokens. \Cref{app:token_level} offers a fine-grained token-level analysis, featuring saliency maps and grid TDVN visualizations. Proofs of \Cref{theorem:bias-var} are given in \Cref{app:proof}, followed by an examination of the i.i.d.\ assumption through repetition experiments in \Cref{app:iid}. We then provide detailed descriptions and illustrations of task-vector accuracy and early-exit accuracy in \Cref{app:tv-early-acc}. Finally, \Cref{app:enhance_tv} presents illustrations of the task-vector contrastive fine-tuning method and PCA visualization of extracted task vectors.

\section{Task Description}
\label{app:tasks}

\subsection{Symbolic Tasks}

Detailed descriptions of the symbolic tasks used in our empirical studies are provided in \Cref{tab:tasks}, covering the algorithmic, translation, linguistic, and knowledge domains.

\begin{table}[h]
    \centering
    \small
    \caption{Descriptions of symbolic tasks.}
    \label{tab:tasks}
    \newcolumntype{T}{r<{$\;$}@{}c<{$\;$}@{}l}  
    \adjustbox{max width=\textwidth}{
    \begin{tabular}{l|lTp{170pt}}
        \toprule
        \textbf{Task Domains} & \textbf{Task} & \multicolumn{3}{c}{\textbf{Example}} & \textbf{Description} \\
        \midrule
        \midrule

        \multirow{10}{*}{Algorithmic(Letter-to-Letter)} & Copy Letter & a & $\to$ & a & Output the same letter of the given letter. \\
        & Next Letter & a & $\to$ & b & Output the next letter of the given letter in the alphabet. \\
        & To Uppercase & a & $\to$ & A & Output the corresponding uppercase letter of the given lowercase letter. \\
        & Prev Letter & b & $\to$ & a & Output the previous letter of the given letter in the alphabet. \\
        & Next 2 Letter & a & $\to$ & c & Output the letter that comes two positions after the given letter in the alphabet. \\
        \midrule
        \midrule

        \multirow{7}{*}{Algorithmic(List-to-Element)} & List First & [a,b,c] & $\to$ & a & Output the first item in the given list. \\
        & List Last & [a,b,c] & $\to$ & c & Output the last item in the given list. \\
        & List Length & [a,b,c] & $\to$ & 3 & Output length of the given list. \\
        & List First Upper & [a,b,c] & $\to$ & A & Get the first item in the given list, then output the corresponding uppercase letter. \\
        & List Last Upper & [a,b,c] & $\to$ & C & Get the last item in the given list, then output the corresponding uppercase letter. \\
        \midrule
        \midrule

        \multirow{5}{*}{Translation} 
         & French $\to$ English  & bonjour & $\to$ & hello     & Translate the given French word into English. \\
         & Spanish $\to$ English & gracias & $\to$ & thank you & Translate the given Spanish word into English. \\
         & English $\to$ French  & goodbye & $\to$ & au revoir & Translate the given English word into French. \\
         & English $\to$ Italian & music   & $\to$ & musica    & Translate the given English word into Italian. \\
         & English $\to$ Spanish & thank you & $\to$ & gracias      & Translate the given English word into Spanish. \\
        \midrule
        \midrule

         \multirow{11}{*}{Linguistic} 
 & Antonyms                   & hot   & $\to$ & cold      & Output the antonym of the given word. \\
 & Plural $\to$ Singular      & cats  & $\to$ & cat       & Convert the given plural noun to its singular form. \\
 & Present Simple $\to$ Gerund& run   & $\to$ & running   & Convert the given verb from present simple to its gerund form. \\
 & Present Simple $\to$ Past Perfect & walk & $\to$ & had walked & Convert the given verb from present simple to past perfect tense. \\
 & Present Simple $\to$ Past Simple  & jump & $\to$ & jumped    & Convert the given verb from present simple to past simple tense. \\
 & Singular $\to$ Plural      & dog   & $\to$ & dogs      & Convert the given singular noun to its plural form. \\

        \midrule
        \midrule
 \multirow{11}{*}{Knowledge}
 & Country $\to$ Capital         & France     & $\to$ & Paris        & Output the capital city of the given country. \\
 & Football Player $\to$ Position& Lionel Messi & $\to$ & Forward     & Output the playing position of the given football player. \\
 & Location $\to$ Continent      & Brazil     & $\to$ & South America & Output the continent where the given location is found. \\
 & Location $\to$ Country        & Kyoto     & $\to$ & Japan        & Output the country in which the given location is situated. \\
 & Location $\to$ Language       & Egypt     & $\to$ & Arabic       & Output the primary language spoken in the given location. \\
 & Location $\to$ Religion       & India     & $\to$ & Hinduism     & Output the predominant religion of the given location. \\
        \bottomrule
    \end{tabular}}
\end{table}

\subsection{Language Understanding Tasks}
For evaluating TDNV on natural language understanding tasks, we require each query sentence to be assessed across multiple attributes. To enable this, we construct a synthetic natural language dataset in which every sentence can be evaluated on six distinct attributes: length, semantic polarity, tense, sentence type, subject person, and entity type. Each attribute is associated with several categorical labels, all of which are summarized in \Cref{tab:languge_attr_label}. The dataset contains 1,000 samples, and representative examples of these sentences are provided in \Cref{tab:language_example}.

\begin{table}[h]
\centering
\adjustbox{max width=0.6\textwidth}{
\begin{tabular}{ll}
\toprule
\textbf{Attribute} & \textbf{Labels} \\
\midrule
Length & short, medium, long \\
Semantic Polarity & positive, negative, neutral \\
Tense & present, past, future, progressive \\
Sentence Type & declarative, interrogative, imperative, exclamatory \\
Subject Person & first\_person, second\_person, third\_person \\
Entity Type & person, location, organization \\
\bottomrule
\end{tabular}
}
\vspace{.1in}
\caption{The attributes and labels in language understanding dataset.}
\vspace{-.3in}
\label{tab:languge_attr_label}
\end{table}

\begin{table}[h]
\centering
\adjustbox{max width=\textwidth}{
\begin{tabular}{p{6cm}ccccccc}
\toprule
\textbf{Sentence} & \textbf{Length} & \textbf{Semantic Polarity} & \textbf{Tense} & \textbf{Sentence Type} & \textbf{Subject Person} & \textbf{Entity Type} \\
\midrule
I enjoy morning walks. & short & positive & present & declarative & first\_person & person \\
Close the window now. & short & neutral & present & imperative & second\_person & location \\
Despite the heavy rain, our research team successfully completed the outdoor experiment and gathered all the required samples before sunset. & long & positive & past & declarative & third\_person & organization \\
Will you be visiting the United Nations headquarters in New York next year to attend the global climate summit? & long & neutral & future & interrogative & second\_person & location \\
While the orchestra rehearsed the challenging new symphony, the conductor meticulously adjusted each section to achieve the perfect balance of sound for the upcoming performance. & long & neutral & progressive & declarative & third\_person & organization \\
\bottomrule
\end{tabular}
}
\vspace{.1in}
\caption{Example sentences in language understanding dataset, each sentence is annotated with 6 attributes.}
\vspace{-.3in}
\label{tab:language_example}
\end{table}

\subsection{Multimodality Tasks}
\begin{figure}[h!]
    \centering
    \includegraphics[width=0.4\linewidth]{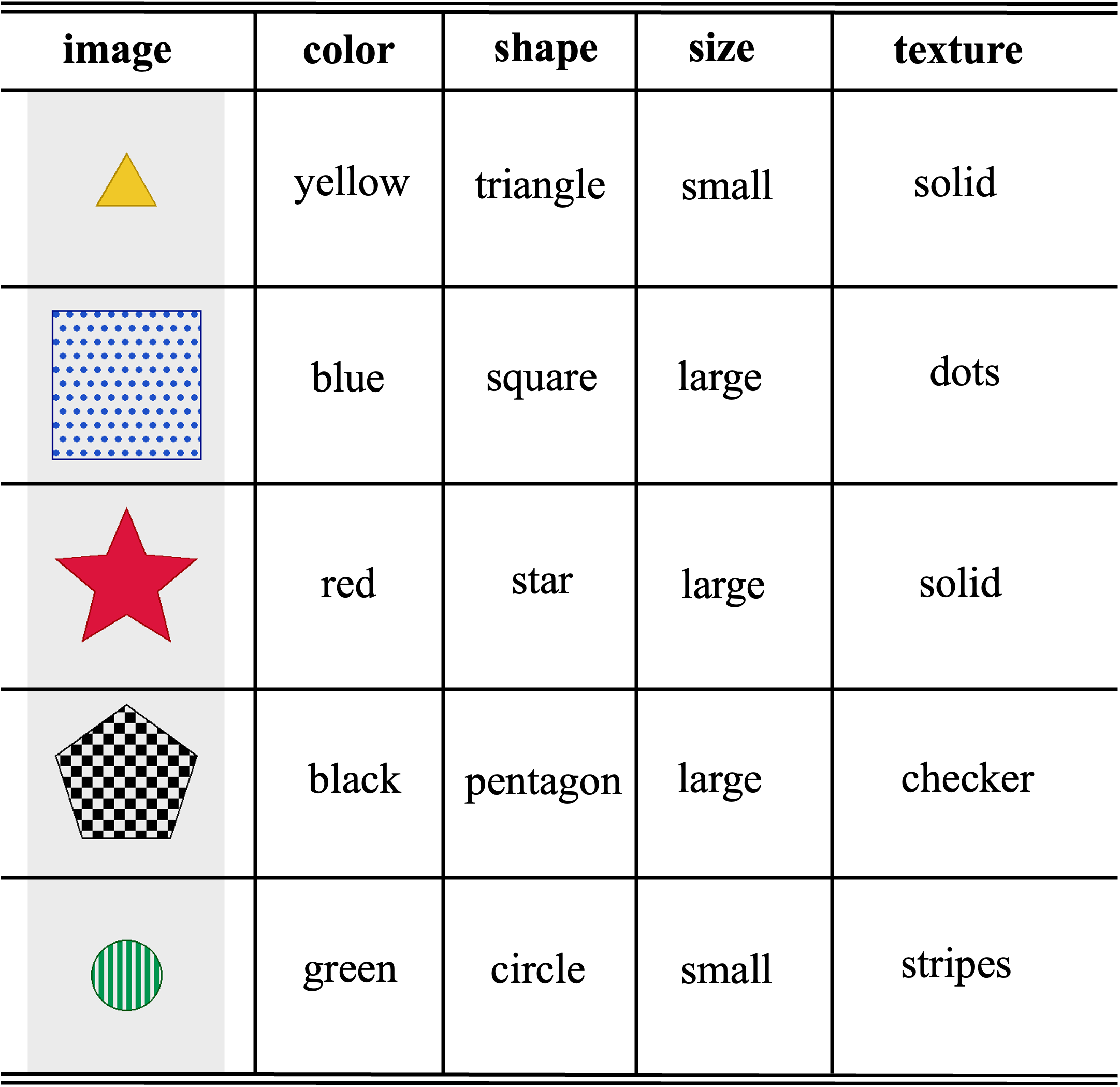}
    \caption{Example images in multimodality dataset, each image is annotated with 4 attributes.}
    \label{fig:mllm_visual}
    \vspace{-.2in}
\end{figure}

For evaluating TDNV on multimodality tasks, we require each query image to be assessed across multiple attributes. To enable this, we construct a synthetic vision-text dataset in which every image can be evaluated on four distinct attributes: color, shape, size and texture. Each attribute is associated with several categorical labels, all of which are summarized in \Cref{tab:mllm_attr_lable}. The dataset contains 300 samples, and representative examples of these images are provided in \Cref{fig:mllm_visual}.

\begin{table}[h]
\centering
\adjustbox{max width=0.5\textwidth}{
\begin{tabular}{ll}
\toprule
\textbf{Attribute} & \textbf{Labels} \\
\midrule
Color   & red, green, blue, yellow, black \\
Shape   & circle, square, triangle, pentagon, star \\
Size    & small, medium, large \\
Texture & solid, stripes, dots, checker \\
\bottomrule
\end{tabular}
}
\vspace{.1in}
\caption{The attributes and labels in multimodality dataset.}
\label{tab:mllm_attr_lable}
\vspace{-.1in}
\end{table}

\section{Related Works}
\label{app:related_works}

\textbf{Layerwise Representations} An intriguing line of research \citep{ben2022nearest, fang2021exploring,  wang2023understanding, rangamani2023feature, he2024law, zhou2025all} has empirically investigated the role of different layers in feature learning. These studies show that in image classification tasks, features in intermediate layers become increasingly linearly separable as the layers deepen. Specifically, Neural Collapse (\NC) properties emerge in intermediate layers, where the within-class variance decreases compared to the between-class variance as depth increases. This indicates that layerwise compression occurs \textbf{monotonically} with layer depth in these settings. However, our hypothesis reveals that in the ICL setting, decoder-only models' layerwise representations exhibit \textbf{dual} encoding-decoding stages: shallow layers compress information while deep layers express it. Furthermore, research by \citep{skean2025layer} shows that intermediate layers consistently outperform both shallow and final layers on downstream tasks. Our research focuses on how intermediate layers achieve maximum information compression in the ICL setting.

\textbf{In-Context Learning Interpretability} Numerous studies have focused on the working mechanisms of ICL \citep{xie2021explanation, chen2021meta, von2023transformers, dai2022can, ahn2023transformers, olsson2022context}. \citet{xie2021explanation} propose that ICL emerges as an implicit form of Bayesian inference. In the realm of meta-learning, \citet{chen2021meta} introduce in-context tuning to predict target labels from concatenated sequences of task instructions. A significant line of research connects ICL with gradient descent optimization. \citet{von2023transformers} demonstrate that Transformers trained on autoregressive tasks can emulate gradient descent in their forward pass. \citet{dai2022can, ahn2023transformers} compare standard gradient descent-based fine-tuning with ICL, revealing that transformer attention in ICL exhibits a dual form of gradient descent-based optimization. \citet{olsson2022context} identify "induction heads"—specific attention heads that facilitate ICL by copying patterns from previous tokens. However, our work focuses on the layer-wise representational analysis of ICL. In addition, \citet{doimo2024representation} also reveals different clustering patterns in shallow and deep layers. However, we observed a strikingly different phenomenon and would like to clarify the differences in settings and results. \citet{doimo2024representation} uses demonstrations with minimal task cues while the query itself provides ample task information, enabling high zero-shot accuracy and only modest few-shot gains. Our tasks encode the task only in the demonstration pairs, while the query provides no clues, yielding near-random zero-shot performance. As a result, our work shows that the model progressively compresses task information from the demonstration pairs, reaching maximal compression around the intermediate layers. In contrast, \citet{doimo2024representation} clusters representations by semantic subject, with ARI peaking in the earliest layers.

\textbf{Task Representations} Researchers have explored various ways to extract compact representations of different ICL tasks, including task vectors \citep{hendel2023context}, function vectors \citep{todd2023function}, and state vectors \citep{li2024context}. Task vectors \citep{hendel2023context} are extracted from the intermediate hidden state of the final separate token. Function vectors \citep{todd2023function} are derived from attention activation through causal mediation, while state vectors \citep{li2024context} concatenate activations from several initial layers of transformers. These representations effectively enable models to perform ICL tasks by injecting to specific transformer layers' hidden states during inference. Some researchers have explored task manipulation within in-context learning. For instance, \citet{shao2023compositional} has demonstrated that compositional task representations can be created through composition model training. Additionally, In-context vectors \citep{liu2023context} enhance ICL through latent space steering. However, previous works have mainly focused on task representations for individual tasks, and none have provided a layer-wise analysis of task vectors. Our research examines how the model distinguishes between different tasks from a geometric perspective across shallow to deep layers.

\section{Additional Experiments}
\label{app:additional_exp}

\subsection{Universal of \phenomenon across different tasks}

In \Cref{fig:main}, we validate the \textit{\phenomenon} phenomenon across layers on Letter-to-Letter tasks. To further assess its generality, we evaluate the phenomenon on the other algorithmic task groups:
(i) List-to-Element tasks (list-first, list-last, list-length, list-first-upper, list-last-upper);
(ii) A combination of Letter-to-Letter and List-to-Element tasks.

\begin{figure}[h]
    \centering
    \begin{subfigure}[t]{0.48\linewidth}
        \centering
        \includegraphics[width=\linewidth]{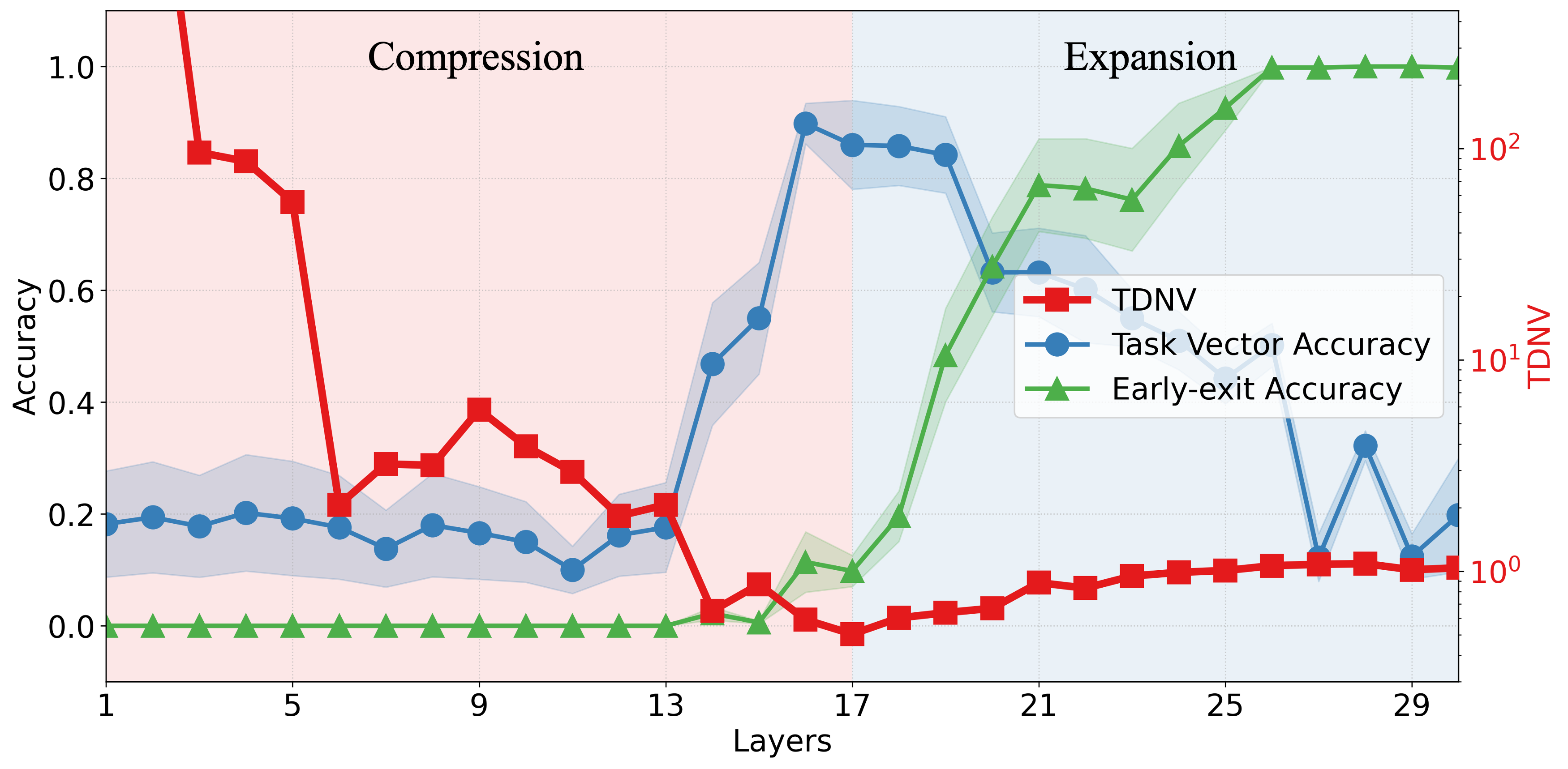}
        \caption{List-to-Element}
        \label{fig:acc_noise}
    \end{subfigure}
    \hfill
    \begin{subfigure}[t]{0.48\linewidth}
        \centering
        \includegraphics[width=\linewidth]{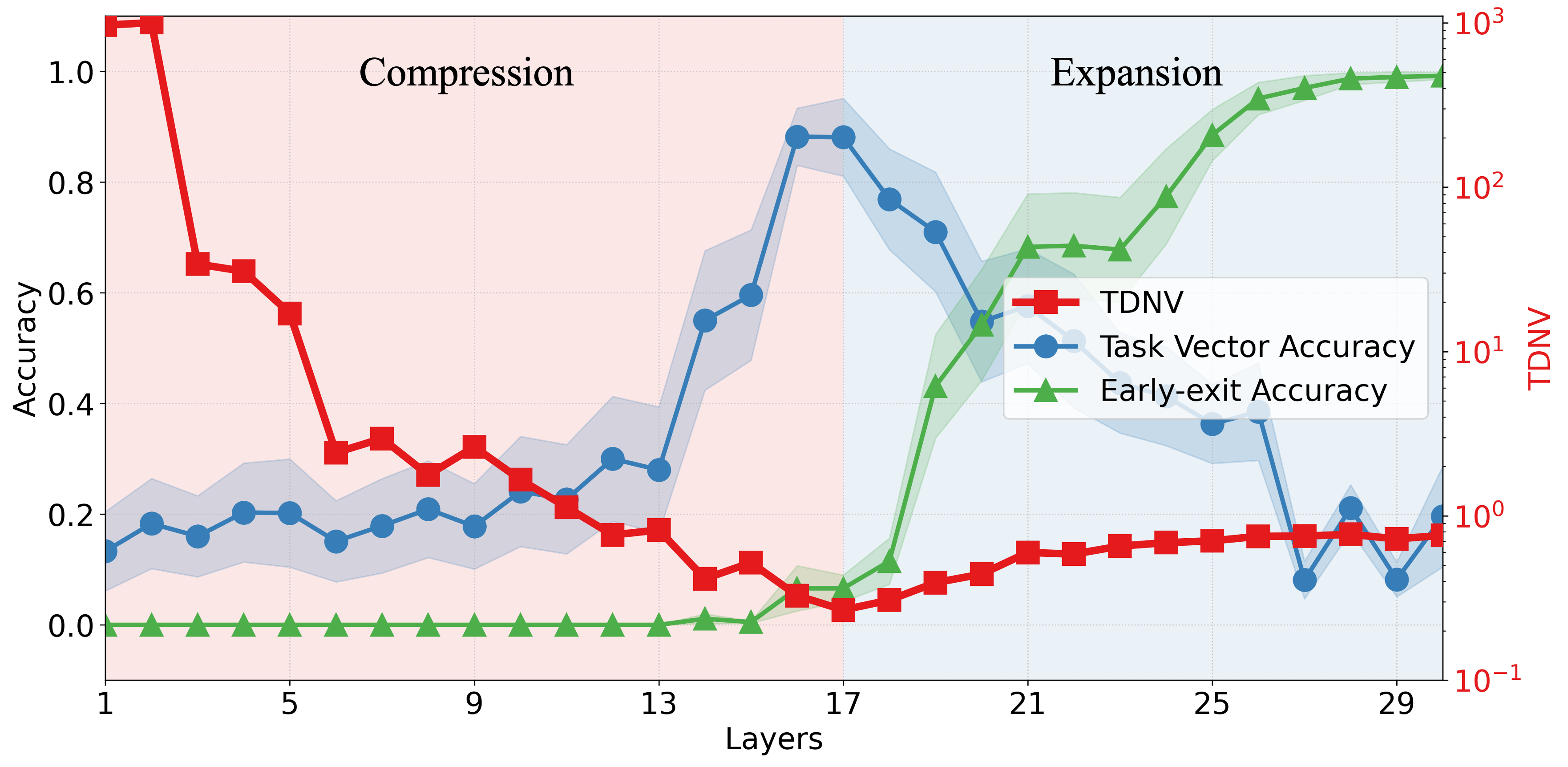}
        \caption{Letter-to-Letter and List-to-Element}
        \label{fig:tdvn_more_tasks}
    \end{subfigure}
   
    \caption{\textit{\phenomenon} phenomenon across different tasks groups: a) List-to-Element, b) a combination of two task groups, the Letter-to-Letter and List-to-Element. TDNV first decreases then increases from shallow to deep layers, splitting the model into compression and expression stages.}
    \label{fig:tdnv_layerwise_tasks}
\end{figure}

We use the DeepSeek-Coder-7B model under a 15-shot ICL setting. As shown in \Cref{fig:tdnv_layerwise_tasks}, the TDNV exhibits a U-shaped curve across layers in both settings, and both task vector accuracy and early-exit accuracy follow patterns similar to those observed in \Cref{fig:main}. Notably, \Cref{fig:tdnv_layerwise_tasks}(b) presents results for the combined task groups listed in \Cref{tab:tasks}, further supporting the conclusion that this phenomenon holds broadly across diverse tasks. These findings further confirm the universality of the \textit{\phenomenon}.

\subsection{Alternative Representations for task vector}
We choose the last separator token as the ICL representation following prior work \citep{hendel2023context}, where it serves as a natural anchor point between the demonstrations and the query. For comparison, we evaluate two other aggregation strategies: ($i$) Mean of All Tokens: the mean of all demonstration token representations(remove the last query and separator), and ($ii$) Mean of All Separator Tokens: the mean of all separator token representations. To quantitatively evaluate which representation best captures layerwise features, we compare the TDNV changes during both compression and expression stages:
\[
\Delta_{\text{Compression}}
= \frac{\mathrm{TDNV}_0 - \mathrm{TDNV}_{\ell_{\min}}}{\mathrm{TDNV}_0}
\qquad\text{and}\qquad
\Delta_{\text{Expression}}
= \frac{\mathrm{TDNV}_L - \mathrm{TDNV}_{\ell_{\min}}}{\mathrm{TDNV}_L}.
\]
Where $\mathrm{TDNV}_0$ and $\mathrm{TDNV}_L$ are the TDNV of the first and last layer, and $\mathrm{TDNV}_{\ell_{\min}}$ is the minimum TDNV.

\begin{table}[ht]
\centering
\begin{tabular}{lcc}
\toprule
\textbf{Representation} & \textbf{Compression Ratio $\uparrow$} & \textbf{Expression Ratio $\uparrow$} \\
\midrule
Mean All Tokens & 0.9671 & 0.6002 \\
Mean Sep Tokens & 0.9879 & 0.5448 \\
\textbf{Last Sep Token} & \textbf{0.9926} & \textbf{0.7746} \\
\bottomrule
\end{tabular}
\caption{Compression and expression ratios for different representations.}
\label{tab:compression_expression}
\end{table}

As shown \Cref{tab:compression_expression}, the Last Sep Token representation demonstrates both higher compression and expression ratios. This evidence supports our conclusion that the last separator token remains the optimal choice for capturing task-relevant information. 

The other two alternatives are suboptimal for the following reasons:
\begin{itemize}
\item Mean of All Tokens. As shown in \Cref{fig:layerwise_saliency}, saliency maps reveal that token contributions to task representation are highly uneven. Early layers focus on label tokens within the demonstrations, while later layers shift attention to the final separator token as the primary carrier of task information. Averaging across all tokens therefore introduces noise from irrelevant content tokens and dilutes the in-context learning (ICL) signal.

\item Mean of All Separator Tokens. As illustrated in \Cref{fig:grid_cdnv_perturb}, grid-level TDNV analysis across separator tokens shows a monotonic decrease in TDNV from the first to the last separator. This pattern indicates that the model progressively compresses task-relevant information across successive demonstrations. Because later separators encode richer context and stronger compression, averaging over all separators weakens this effect, whereas using only the final separator captures the fully accumulated task representation.
\end{itemize}

\section{Token Level Analysis}
\label{app:token_level}

\subsection{Saliency Maps}
\label{app:saliency}

In the main pages, we have extensively explored the use of TDNV to quantify layerwise information compression during ICL, revealing important statistical properties of model representations. In this subsection, we complement the understanding from a fine-grained token level. In particular, we will use the method of saliency maps \citep{wang2023label}, specifically elucidating which parts of the input significantly contribute to the model's decision-making. By highlighting critical token interactions, saliency maps provide intuitive insights into model behavior. Denoting by $I_\ell$  saliency map at the $\ell$-th layer, we compute it by,
\begin{equation}
I_\ell = \left| \sum_h A_{h,\ell} \odot \frac{\partial \mathcal{L}}{\partial A_{h,\ell}} \right|,
\end{equation}
where $A_{h,\ell}$ represents the attention map of head $h$ at layer $\ell$, and the loss $\mathcal{L}$ is the cross-entropy calculated between the logits of the last token and the ground-truth label. Thus, a saliency map quantifies the importance of internal activations by integrating both attention strength and its gradient-based influence on the model's outcome. In a nutshell, these maps highlight how token interactions evolve across layers.  

\begin{figure}[t]
    \centering
    \includegraphics[width=1\linewidth]{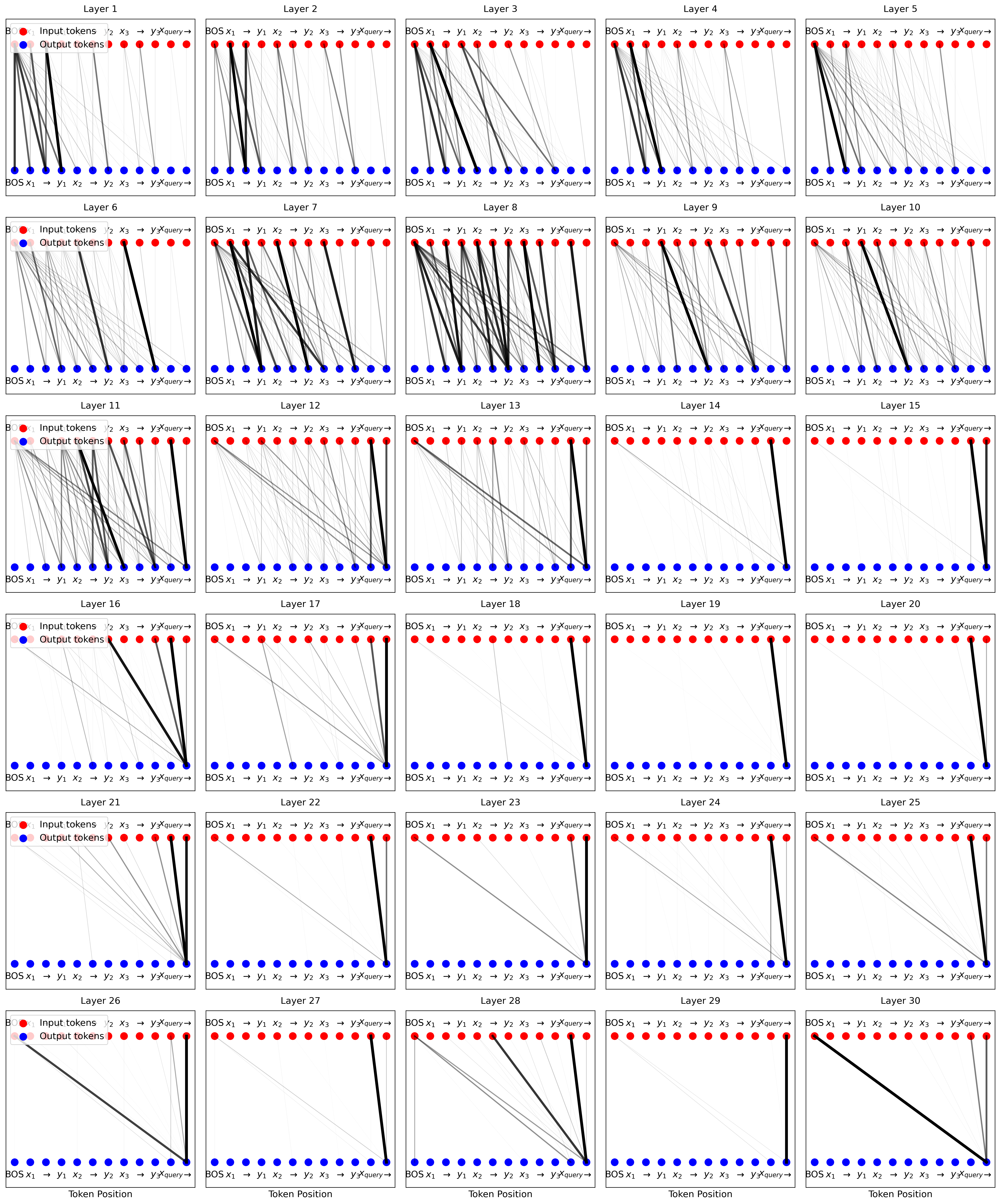}
    \caption{Saliency Maps for all layers.}
    \label{fig:saliency-all layers}
\end{figure}

We show saliency maps of all layers using three demonstrations in \Cref{fig:saliency-all layers}. In shallow layers, there is more interaction within demonstrations, indicating that the model extracts task information from each demonstration. In deep layers, there is less interaction within demonstrations and more interaction with the last token, indicating that the model uses the accumulated task information to generate output.

\begin{figure}[t]
    \centering
    \begin{subfigure}[t]{0.32\linewidth}
        \centering
        \includegraphics[width=\linewidth]{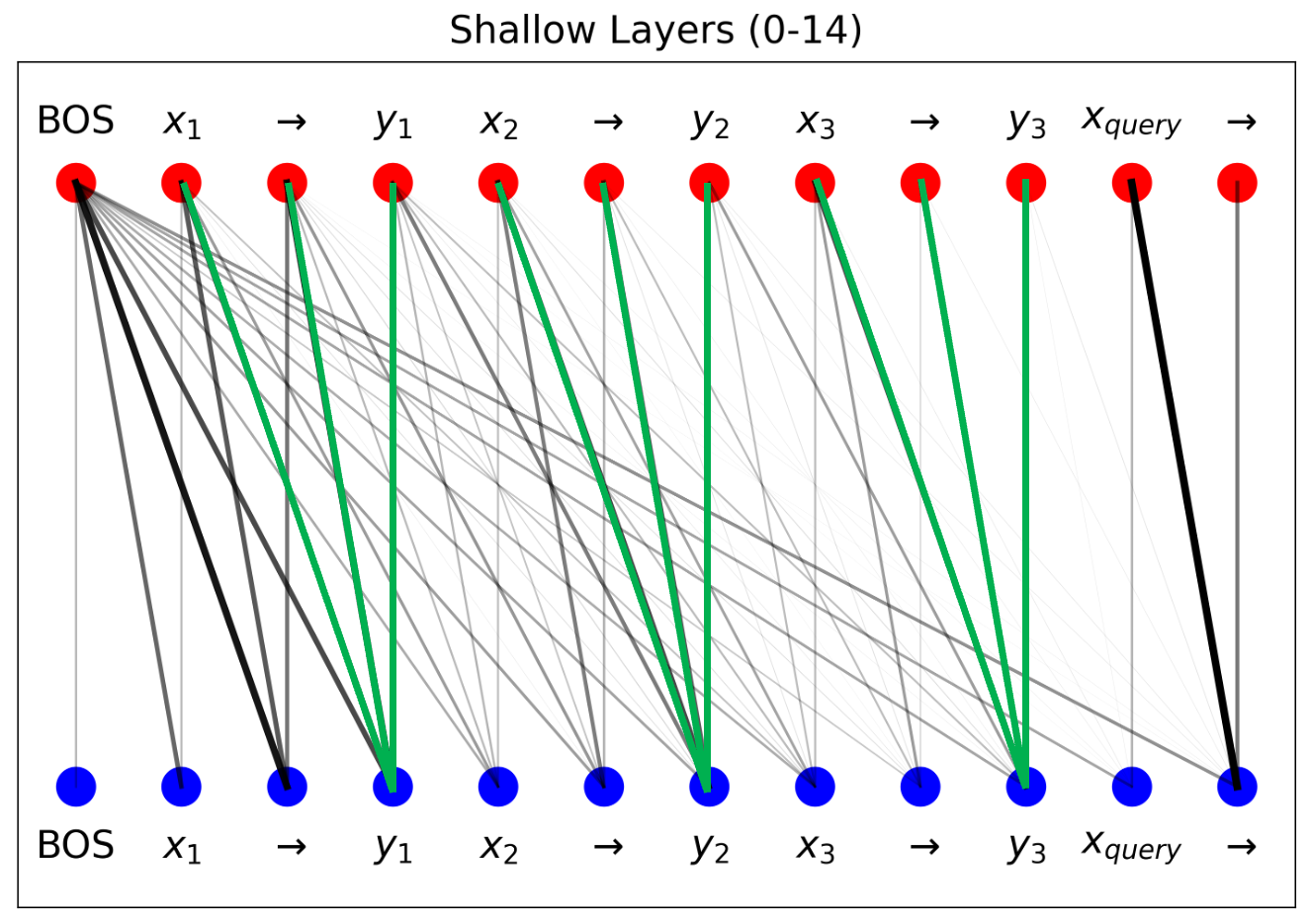}
        \label{fig:shallow_saliency}
    \end{subfigure}
    \hfill
    \begin{subfigure}[t]{0.32\linewidth}
        \centering
        \includegraphics[width=\linewidth]{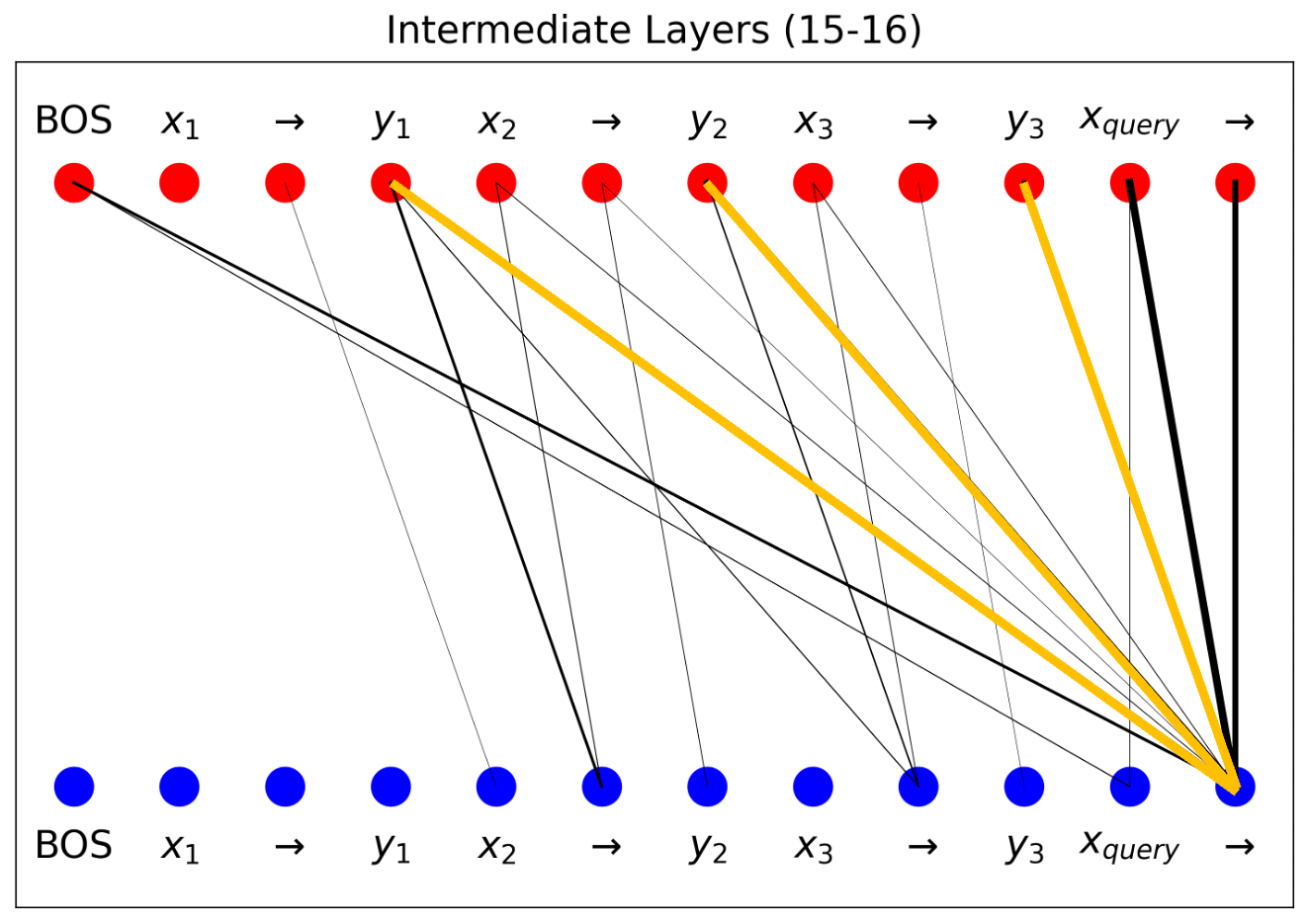}
        \label{fig:intermediate_saliency}
    \end{subfigure}
    \hfill
    \begin{subfigure}[t]{0.32\linewidth}
        \centering
        \includegraphics[width=\linewidth]{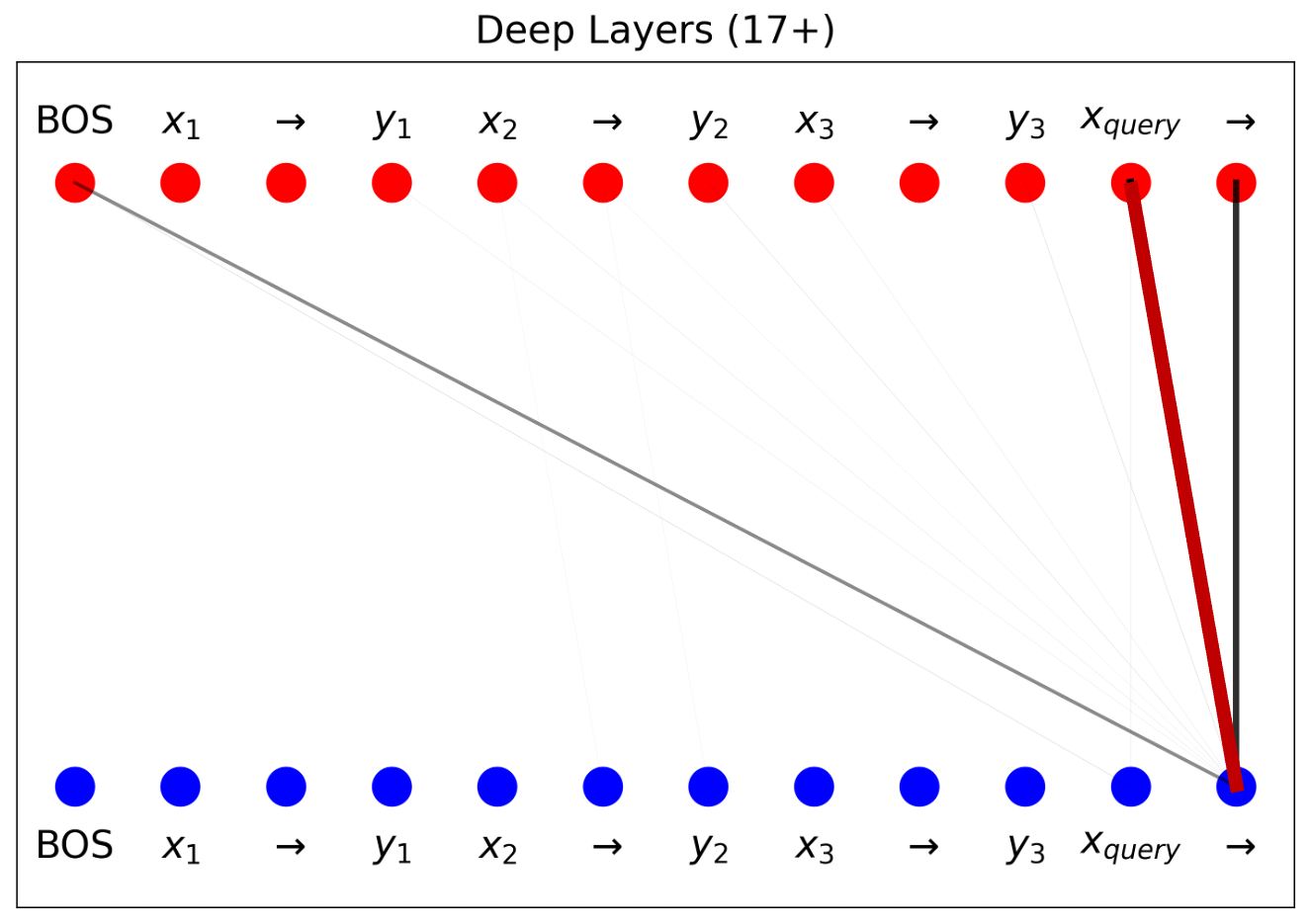}
        \label{fig:deep_saliency}
    \end{subfigure}
    \caption{Saliency maps across transformer layers: (a) shallow, (b) intermediate, and (c) deep. The edge widths indicate saliency magnitude from input tokens (red dots) to output tokens (blue dots).}
    \label{fig:layerwise_saliency}
\end{figure}

To observe more clear patterns, we group the layers into 3 groups: shallow (average of layers 1-14), intermediate (average of layers 15-16), and deep (layers 17 and above) in \Cref{fig:layerwise_saliency}. We observe that $(i)$ in shallow layers, the model focuses on token-to-label mappings within demonstrations, reflecting local compression, $(ii)$ intermediate layers shift focus toward the final token, indicating integration of task-level information, and $(iii)$ deep layers emphasize interactions between the query and final token, aligning with expression and prediction. Thus, token-level interpretability also aligns with the layerwise compression-expression trajectory. 

\subsection{Grid-TDNV}
\label{app:grid_tdnv}
\begin{figure}[t]
    \centering
    \includegraphics[width=1\linewidth]{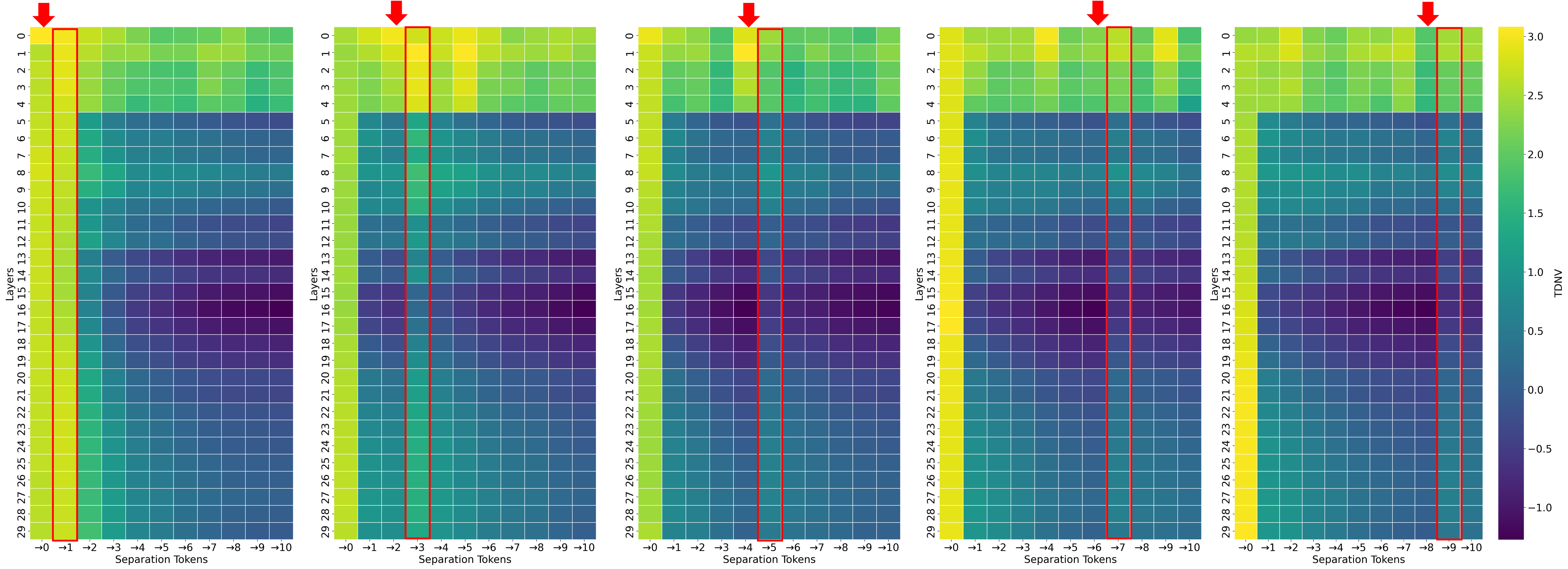}
    \caption{The grid TDNV pattern when perturbing one demonstration at different positions: from left to right,  the perturbation is applied to the 0th, 2nd, 4th, 6th, and 8th demonstration.}
    \label{fig:grid_cdnv_perturb}
\end{figure}

In \Cref{sec:robustness}, we find that the position of noisy demonstration affects compression. To investigate why perturbing later demonstrations results in higher TDNV at the final separation token, we conduct a more fine-grained token-level analysis by computing the layerwise TDNV for each separation token across all demonstrations. For each demonstration’s separator, we calculate the layerwise TDNV and organize them into a grid structure, referred to as the grid TDNV. As shown in \Cref{fig:grid_cdnv_perturb}, perturbing a demonstration at a given position most significantly increases the TDNV of the next demonstration. However, this increase gradually diminishes as more correct demonstrations are appended. This pattern suggests that the negative impact of early errors can be partially mitigated by subsequent correct examples.

\section{Proof of \Cref{theorem:bias-var}}
\label{app:proof}

\begin{proof}
\textbf{(Proof for Variance Decay.)} Consider one layer linear attention ~\citep{von2023transformers,ahn2023transformers,wang2024transformers,li2024context} that preserves the normalization property of softmax attention\citep{katharopoulos2020transformers, shen2021efficient}: the hidden state of the query token becomes
\begin{equation}
\begin{aligned}
    \vh'_{q}(K) &= [ \mathrm{Attn}(\vh_1, \dots, \vh_K, \vh_{q})]_{:, K+1} \\
    &= \sum_{i=1}^K \frac{\phi(\vq_{q})^\top \phi(\vk_i)}{K+1 } \vv_{i} +  \frac{\phi(\vq_{q})^\top \phi(\vk_{q})}{K+1} \vv_{q},
\end{aligned}
\end{equation}
where $\vq =\mW^{Q}\vh, \vk =\mW^{K}\vh,\vv =\mW^{V}\vh$ are the query, key and value vectors, respectively, and $\phi : \mathbb{R}^d \to \mathbb{R}^r$ denotes the feature map that approximates the softmax. Define $z_j := \phi(\vq_{q})^\top \phi(\vk_j)$, $z_{q} := \phi(\vq_{q})^\top \phi(\vk_{q})$, we can rewrite it as

\begin{equation}
    \vh'_{q}(K) = \sum_{i=1}^{K} \frac{z_i}{K + 1} \, \vv_i + \frac{z_{q}}{K + 1} \, \vv_{q}.
\label{eq:decom-h}\end{equation}

To further simplify the notation, we define $\va_i:= z_i \vv_i, \quad i = 1, \dots, K, $ and  $\va_{K+1} := z_{q} \vv_{q}$, which gives
\begin{equation}
     \vh'_{q}(K) = \frac{1}{K+1} \sum_{i = 1}^{K+1} \va_i.
\end{equation}

Since $\va_i := \phi(\vq_{q})^\top \phi(\mW^{K}\vh_i) \mW^{V}\vh_i$ only relates to $\vh_i$ and  $\vh_i$ are i.i.d., the $\va_i$ are also i.i.d. with

\[
\mathbb{E}[\va_i] = \boldsymbol{\mu}_a, \qquad
\mathrm{Cov}(\va_i) = \boldsymbol{\Sigma}_a.
\]
Finally, we get,
\begin{equation}
    \mathrm{Var}(\|\vh'_{q}\|_2^2) = \frac{1}{(K+1)^2} \sum_{i=1}^{K+1} \mathrm{Var}(\|\va_i\|_2^2) 
= \frac{K+1}{(K+1)^2} \mathrm{Tr}(\boldsymbol{\Sigma}_a)
= \frac{\mathrm{Tr}(\boldsymbol{\Sigma}_a)}{K+1} 
\sim \mc O(1/K).
\end{equation}

\end{proof}

\begin{proof}
\textbf{(Proof for Mean Shift.)}  Define the mean of demonstrations and the zero-shot as
\[
\boldsymbol{\mu}_{\text{zero-shot}} := \mathbb{E}[z_{q} \vv_{q}] , \qquad \boldsymbol{\mu}_{\text{demo}} := \mathbb{E}[z_i \vv_i].
\]
Then according to \eqref{eq:decom-h}, we have
\begin{equation}
    \mathbb{E}[h'_{q}(K)] = 
\frac{1}{K + 1} \underbrace{\mathbb{E}[z_{q} \vv_{q}]}_{=: \boldsymbol{\mu}_{\text{zero-shot}}}
+ 
\frac{K}{K + 1} \underbrace{\mathbb{E}[z_i \vv_i]}_{=: \boldsymbol{\mu}_{\text{demo}}}.
\end{equation}

If $K = 0$, we get $
    \mathbb{E}[\vh'_{q}(0)] = \mathbb{E}[z_{q} \vv_{q}]$. When $K \to \infty$, we have $
    \mathbb{E}[\vh'_{q}(\infty)] = \mathbb{E}[z_i \vv_i]$.

In summary, we get,

\begin{equation}
     \mathbb{E}[\vh_{q}'(K)] = \lambda_K\, \mathbb{E}[\vh_{q}'(0)] + (1 - \lambda_K)\, \mathbb{E}[\vh_{q}'(\infty)]
\end{equation}
where $\lambda_K = 1/(1+K)\sim \mc O(1/K)$.

\end{proof}

\section{Validation of the I.I.D. Assumption under Repitition}
\label{app:iid}
In \Cref{theorem:bias-var}, we assume that each demonstration $\vh_i,i =1,\ldots,K$ is i.i.d. randomly generated from a distribution $\calH$ on $\R^d$. To validate the importance of this assumption, we design experiments that increase the ICL length by providing additional demonstrations as input. There are two possible extension methods:

\begin{itemize}
[leftmargin=12pt,itemsep=2pt,topsep=0pt,parsep=0pt]
\item Repeat Mode: Extending demonstrations by repeating existing examples.
\item Distinct Mode: Extending demonstrations by adding new, unique examples.
\end{itemize}

In the Distinct Mode, new demonstrations are independently and identically distributed (i.i.d.), randomly generated from a distribution. In the Repeat Mode, demonstrations are no longer i.i.d. We examine both performance and compression level across these two modes.

\begin{figure}[h]
    \centering
    \begin{minipage}[t]{0.48\linewidth}
        \centering
        \includegraphics[width=\linewidth]{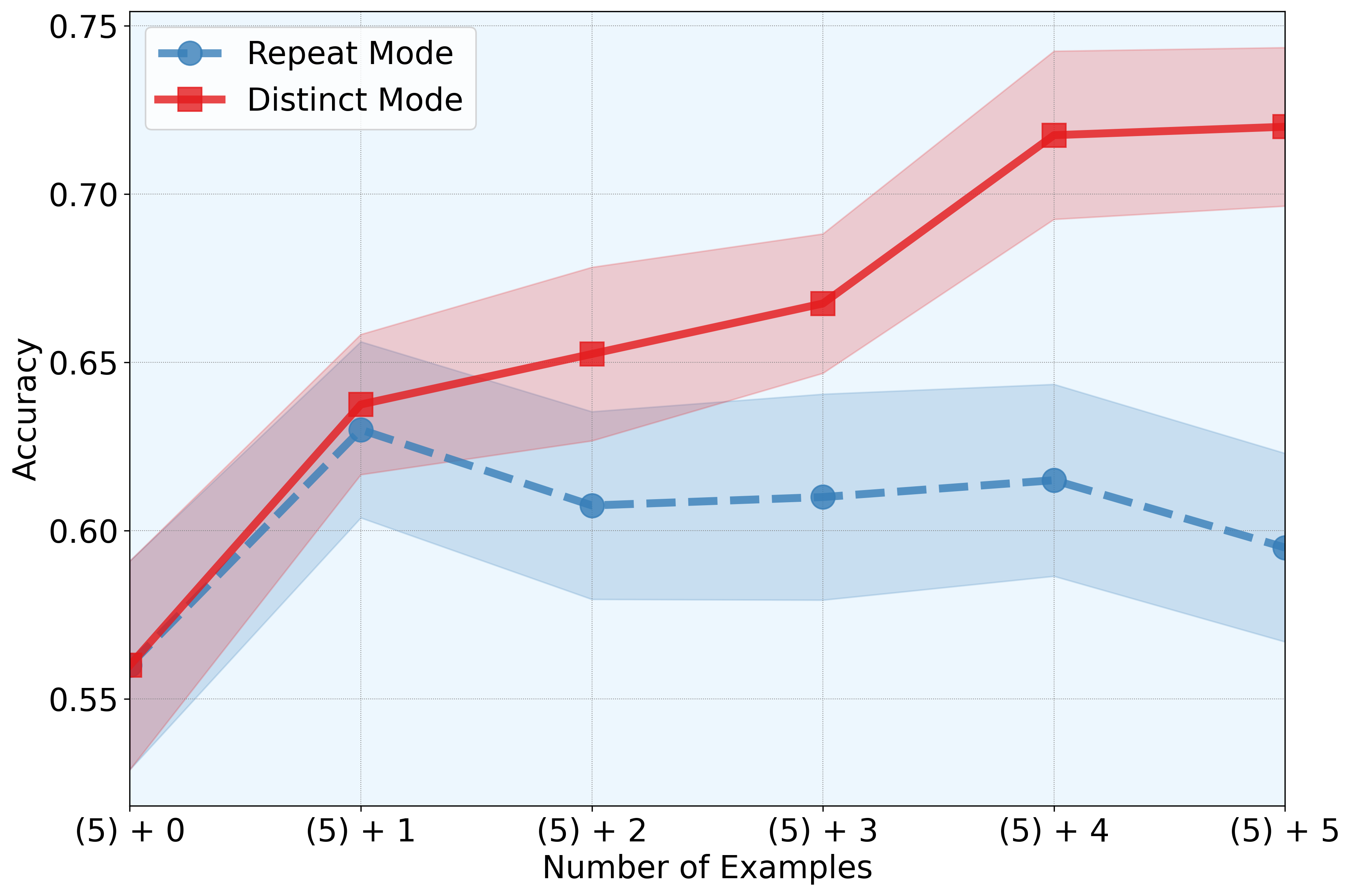}
        \caption{Comparison of accuracy under repeat mode v.s. distinct mode.}
        \label{fig:repeat_performance}
    \end{minipage}
    \hfill
    \begin{minipage}[t]{0.48\linewidth}
        \centering
        \includegraphics[width=\linewidth]{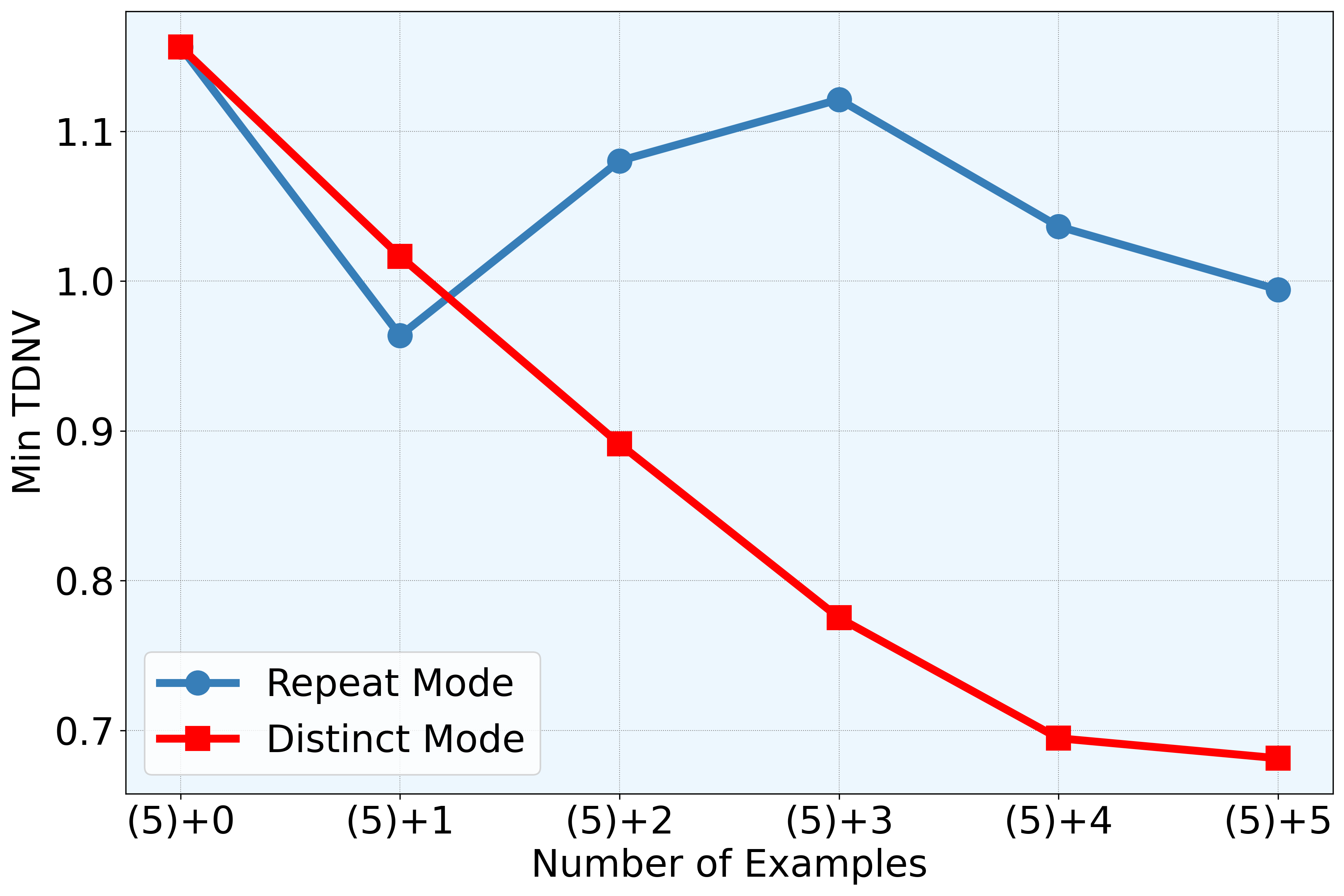}
        \caption{Minimum TDNV under under repeat mode v.s. distinct mode.}
        \label{fig:repeat_cdnv}
    \end{minipage}
\end{figure}
As shown in \Cref{fig:repeat_performance} and \Cref{fig:repeat_cdnv}, in distinct mode, as new demonstrations introduce novel information, the accuracy increases and TDNV decreases. However, in repeat mode, since repeated demonstrations add no new task information, the accuracy and TDNV remain largely unchanged. This proves that when the i.i.d. condition is violated, increasing the number of demonstrations does not lead to better performance and compression.

\section{Task vector accuracy \& Early-exit accuracy}
\label{app:tv-early-acc}

\begin{figure}[h]
    \centering
    \includegraphics[width=0.75\linewidth]{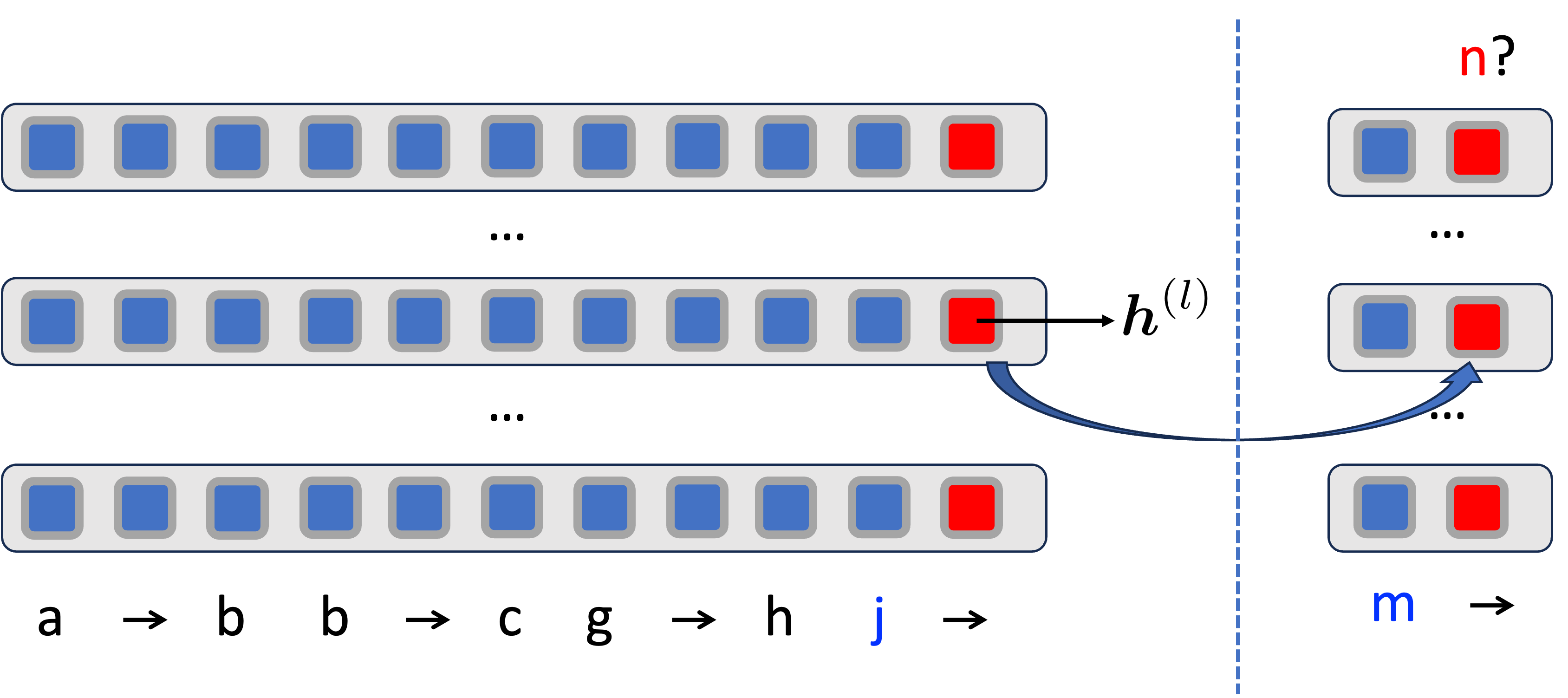}
    \caption{Illustration for task vector accuracy.}
    \label{fig:tv-acc}
\end{figure}

Task vector accuracy\citep{hendel2023context} refers to how accurately a LLM can perform a task using only a learned representation of the task(task vector), instead of full in-context demonstrations. As shown in \Cref{fig:tv-acc}, to evaluate the task vector accuracy, we conduct the following steps:

\begin{enumerate}
[leftmargin=12pt,itemsep=2pt,topsep=0pt,parsep=0pt]
    \item \textbf{Extract} task vector $\boldsymbol{\theta}$ from the demonstration set $\mc S_K$ using a dummy query $x'$, avoiding information leakage from the real query as,
    \begin{equation}
        \boldsymbol{\theta}_{\text{task}} = [f_{\theta^{(1:\ell)}}([\mathcal{S}_K, x'])]_{:,K+1}
    \end{equation}
    \item \textbf{Inject} $\boldsymbol{\theta}$ into a forward pass of the model with only the query $x$, not the full demonstration set, then predict the output using this modulated forward pass.
    \begin{equation}
        y = f_{\theta}([x]; \boldsymbol{\theta}_{task})
    \end{equation}
    
\end{enumerate}

where $[]_{:,K+1}$ means the $(K+1)$-th column, $\ell$ is the layer with most compression. 

\begin{figure}[t]
    \centering
    \includegraphics[width=0.77\linewidth]{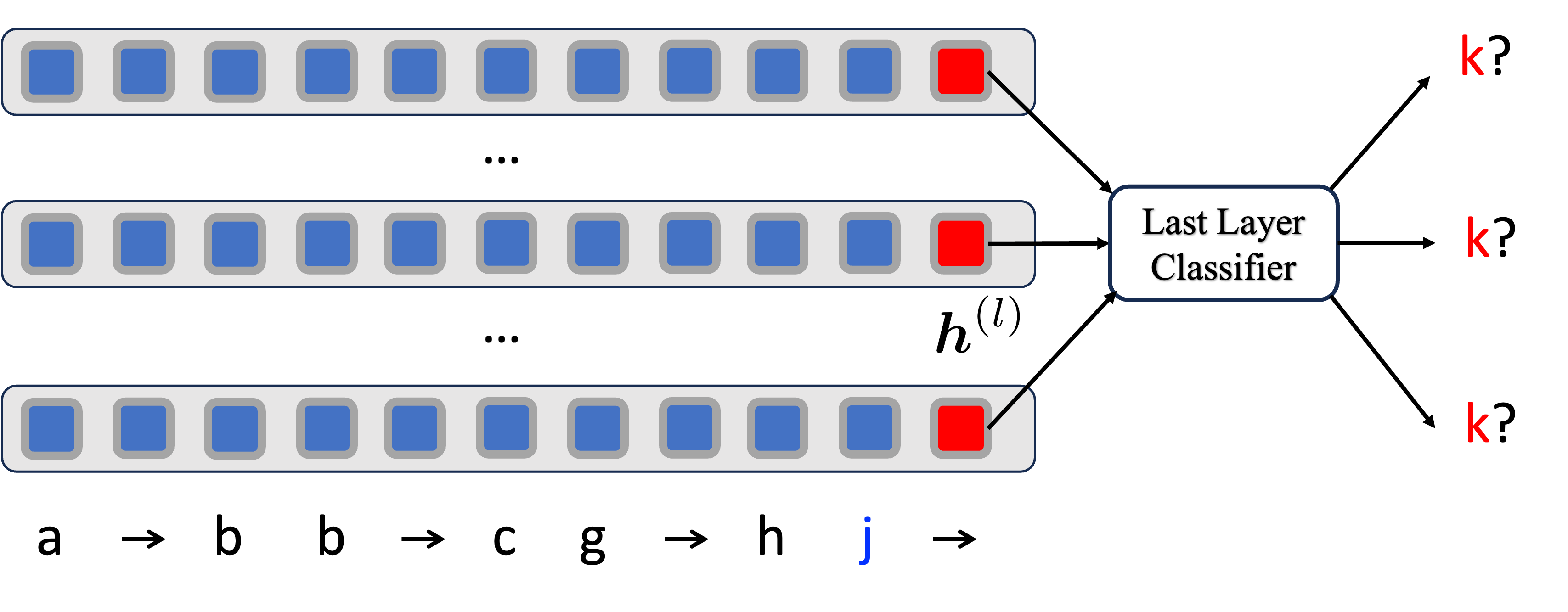}
    \caption{Illustration for early-exit accuracy.}
    \label{fig:early-exit}
\end{figure}

Early-exit accuracy \citep{xin2020deebert, jiang2024layer} measures a model's prediction accuracy when using intermediate-layer representations instead of the final layer for making predictions. This metric helps assess how effectively different layers encode the information needed to complete the task. Let $\vh^{(\ell)} \in \mathbb{R}^d$ be the hidden state of the final token at layer $\ell$, and let $\mc C : \mathbb{R}^d \rightarrow \mathbb{R}^V$ be the last-layer classifier mapping from hidden dimension $d$ to vocabulary size $V$. Then as shown in \Cref{fig:early-exit}, the prediction is,

\begin{equation}
    \hat{y}^{(\ell)} = \arg\max_{v \in \mathcal{V}} \; \text{softmax}(\mc C(\mathbf{h}^{(\ell)}))_v
\end{equation}

Then, the early-exit accuracy at layer $\ell$ over $N$ examples is,
\begin{equation}
    \text{Acc}^{(\ell)} = \frac{1}{N} \sum_{i=1}^{N} \mathbf{1} \left[ \hat{y}_i^{(\ell)} = y_i \right]
\end{equation}

\section{Task-Vector Contrastive Fine-tuning}
\label{app:enhance_tv}
\begin{figure}[h]
    \centering
    \includegraphics[width=0.75\linewidth]{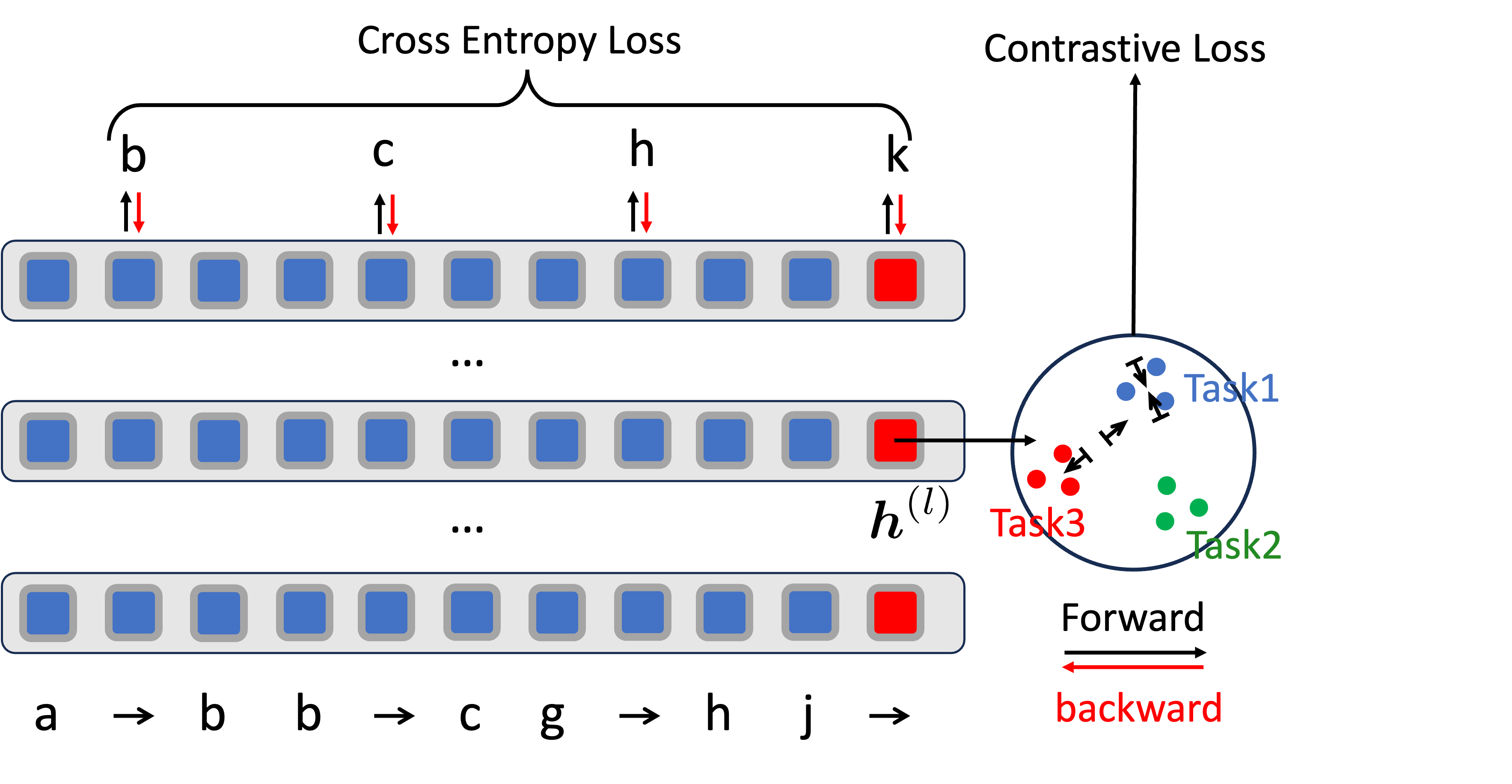}
    \caption{Illustration of task-vector contrastive fine-tuning.}
    \label{fig:contrastive}
\end{figure}

As shown in \Cref{fig:contrastive}, to obtain more compressed and discriminative task vectors, we add an additional contrastive loss on task vectors that explicitly encourages representation compression. Specifically, we fine-tuning a pretrained GPT-2 model on algorithmic ICL tasks domains (next letter, next two letters, previous letter, uppercase, next uppercase letter) using a combined objective as,
\begin{equation}
\mathcal{L}=\mathcal{L}_{\mathrm{CE}}+\beta \mathcal{L}_{\mathrm{Contra}},
\end{equation}
where $\mathcal{L}_{\mathrm{CE}}$ is the cross-entropy loss computed only on separator tokens to predict the correct label and $\mathcal{L}_{\mathrm{Contra}}$ is a contrastive term that pulls hidden states from the same task closer while pushing those from different tasks apart:
\begin{equation}
\mathcal{L}_{\mathrm{Contra}}
= -\frac{1}{|\Omega|}
\sum_{t}\sum_{\substack{i,j\\ i\neq j}}
\log
\frac{\exp(\mathbf{h}_{i,t}^{\top}\mathbf{h}_{j,t}/\tau)}
{\sum_{\substack{(a,b)\neq(i,t)}}
\exp(\mathbf{h}_{i,t}^{\top}\mathbf{h}_{a,b}/\tau)} .
\end{equation}

where $\vh_{i,t}$ denotes the normalized hidden state of the last token in the intermediate layer(task vector) for the $i$-th sample of task $t$, $\tau$ is the temperature, and $\Omega$ is the set of all pairs. We set $\beta=0.1$, temperature $\tau=0.07$, and batch size $100$ with equal samples from each task. Each training example provides $K=20$ context demonstrations followed by a separator token. We finetuned the model to achieve 100\% ICL accuracy on all tasks. 

For evaluation, we extract task vectors with a context length of
$K=20$ and assess task-vector accuracy: performing zero-shot ICL with the extracted task vector injected at the same position. Higher task-vector accuracy reflects more effective task vectors. As shown in \Cref{fig:tv_acc_gpt2}, task-vector contrastive fine-tuning produces better task vectors than standard fine-tuning.

\begin{figure}[h]
    \centering
    \begin{subfigure}[b]{0.48\linewidth}
        \centering
        \includegraphics[width=\linewidth]{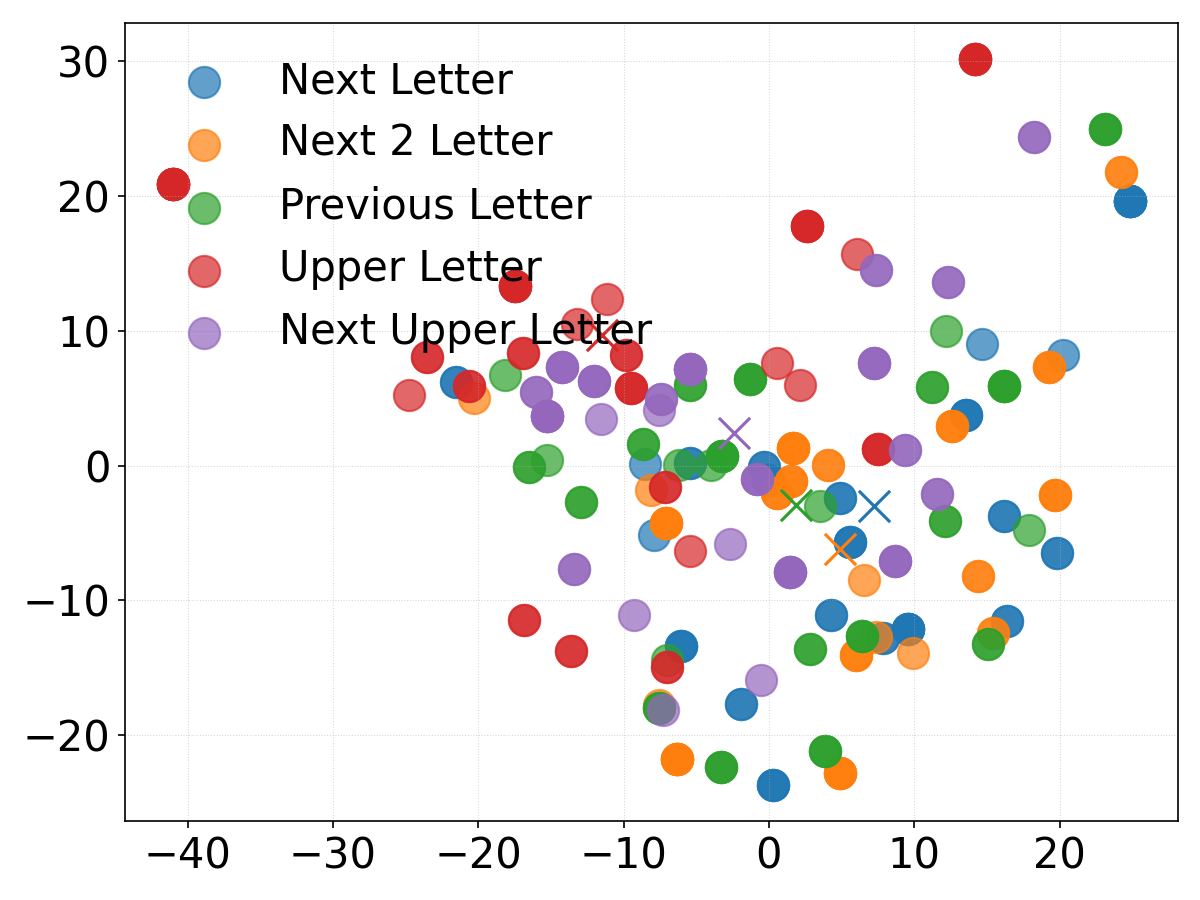}
        \caption{CE loss.}
        \label{fig:pca_baseline_L6}
    \end{subfigure}
    \hfill
    \begin{subfigure}[b]{0.48\linewidth}
        \centering
        \includegraphics[width=\linewidth]{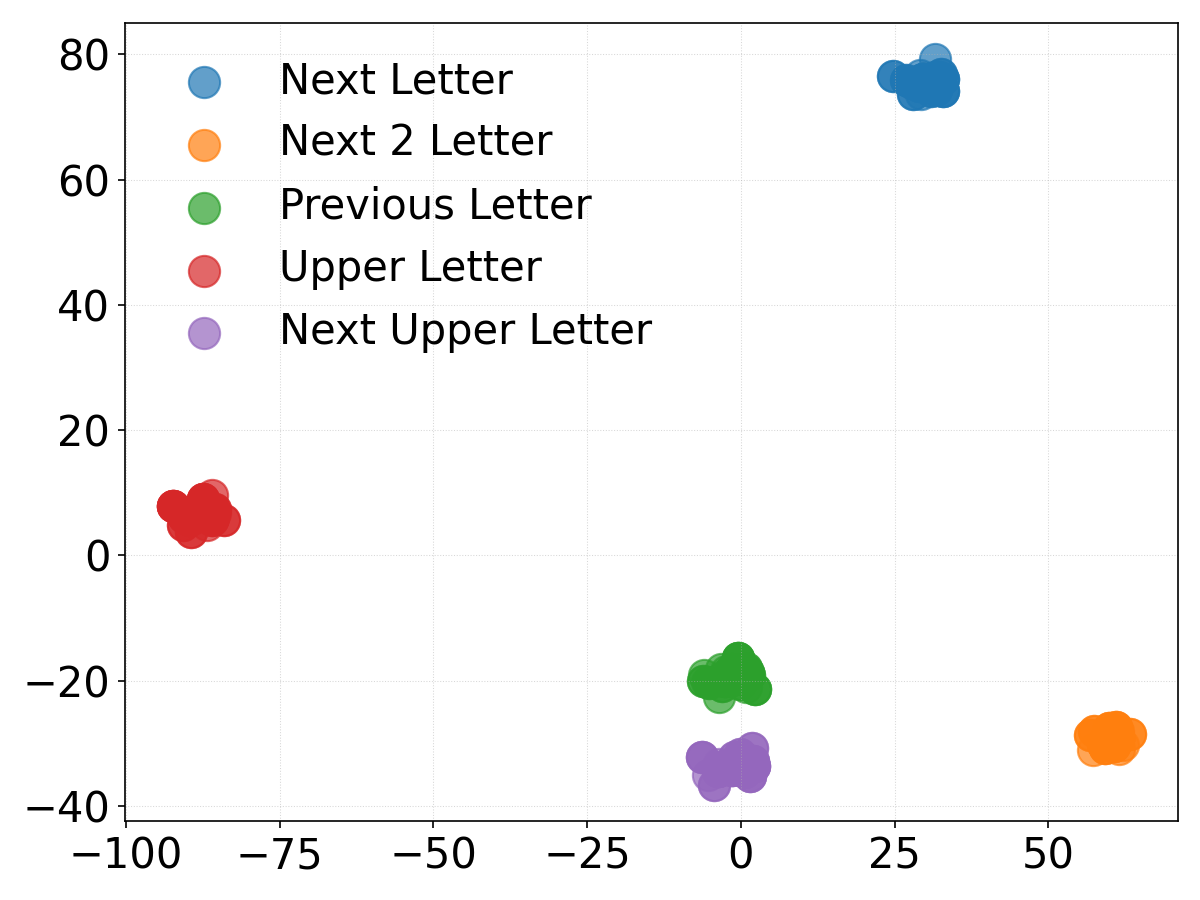}
        \caption{CE loss + contrastive loss.}
        \label{fig:pca_contrast_L6}
    \end{subfigure}
    \caption{Comparison of PCA visualizations for task vectors. The task vector is extracted from models finetuned with different losses. }
    \label{fig:pca_gpt2_comparison}
\end{figure}

To illustrate whether task-vector contrastive fine-tuning produces more compressed task vectors, we visualize hidden states from the $7$-th layer using PCA. As shown in \Cref{fig:pca_gpt2_comparison}, the left panel depicts task vectors extracted from model finetuned with only the cross-entropy loss, while the right panel shows vectors extracted from model finetuned with both cross-entropy and contrastive losses. The latter exhibits more distinct and tightly clustered task representations, demonstrating the effectiveness of the contrastive objective in compressing task vectors. 

\end{document}